%% file: conference_101719.tex
\documentclass[conference]{IEEEtran}
\IEEEoverridecommandlockouts
\usepackage{cite}
\usepackage{amsmath,amssymb,amsfonts}
\usepackage{algorithmic}
\usepackage{graphicx}
\usepackage{textcomp}
\usepackage{xcolor}
\usepackage{authblk}
\usepackage{booktabs}
\usepackage{makecell}

\def\BibTeX{{\rm B\kern-.05em{\sc i\kern-.025em b}\kern-.08em
    T\kern-.1667em\lower.7ex\hbox{E}\kern-.125emX}}
\begin{document}

\title{The Strong Pull of Prior Knowledge \\ in Large Language Models and \\ Its Impact on Emotion Recognition
\thanks{Funded in part by DARPA under contract HR001121C0168}
}

\makeatletter
\newcommand\email[2][]%
   {\newaffiltrue\let\AB@blk@and\AB@pand
      \if\relax#1\relax\def\AB@note{\AB@thenote}\else\def\AB@note{\relax}%
        \setcounter{Maxaffil}{0}\fi
      \begingroup
        \let\protect\@unexpandable@protect
        \def\thanks{\protect\thanks}\def\footnote{\protect\footnote}%
        \@temptokena=\expandafter{\AB@authors}%
        {\def\\{\protect\\\protect\Affilfont}\xdef\AB@temp{#2}}%
         \xdef\AB@authors{\the\@temptokena\AB@las\AB@au@str
         \protect\\[\affilsep]\protect\Affilfont\AB@temp}%
         \gdef\AB@las{}\gdef\AB@au@str{}%
        {\def\\{, \ignorespaces}\xdef\AB@temp{#2}}%
        \@temptokena=\expandafter{\AB@affillist}%
        \xdef\AB@affillist{\the\@temptokena \AB@affilsep
          \AB@affilnote{}\protect\Affilfont\AB@temp}%
      \endgroup
       \let\AB@affilsep\AB@affilsepx
}
\makeatother

\author[1]{Georgios Chochlakis}
\author[1,2]{Alexandros Potamianos}
\author[1]{Kristina Lerman} 
\author[1]{Shrikanth Narayanan}

\affil[ ]{
\textit{chochlak@usc.edu},
\textit{potam@central.ntua.gr},
\textit{lerman@isi.edu},
\textit{shri@ee.usc.edu}
}

\affil[ ]{}
\affil[1]{University of Southern California}
\affil[2]{National Technical University of Athens}

\maketitle

\begin{abstract}
In-context Learning (ICL) has emerged as a powerful paradigm for performing natural language tasks with Large Language Models (LLM) without updating the models' parameters, in contrast to the traditional gradient-based finetuning. The promise of ICL is that the LLM can adapt to perform the present task at a competitive or state-of-the-art level at a fraction of the cost. The ability of LLMs to perform tasks in this few-shot manner relies on their background knowledge of the task (or \textit{task priors}).
However, recent work has found that, unlike traditional learning, LLMs are unable to fully integrate information from demonstrations that contrast task priors. This can lead to performance saturation at suboptimal levels, especially for subjective tasks such as emotion recognition, where the mapping from text to emotions can differ widely due to variability in human annotations.
In this work, we design experiments and propose measurements to explicitly quantify the consistency of proxies of LLM priors and their pull on the posteriors. We show that LLMs have strong yet inconsistent priors in emotion recognition that ossify their predictions. We also find that the larger the model, the stronger these effects become.
Our results suggest that caution is needed when using ICL with larger LLMs for affect-centered tasks outside their pre-training domain and when interpreting ICL results.
\end{abstract}


\section{Introduction}

Large Language Models (LLMs) \cite{radford2019language, ouyangTrainingLanguageModels2022, touvronLlamaOpenFoundation2023, Touvron2023a, zheng2023judging, brownLanguageModelsAre2020, achiam2023gpt} have revolutionized Natural Language Processing, and are now being used as reliable, affordable, and scalable tools to analyze big data in many disciplines~\cite{ziems2023can, kim2024health}, including policy-making~\cite{bakker2022fine}. The ease of use of LLMs comes from their emergent ability to perform downstream tasks by conditioning on input-output demonstrations and/or task instructions, an ability referred to as In-Context Learning (ICL) \cite{wei2022emergent, brownLanguageModelsAre2020}.

While ICL is often contrasted with in-weights learning \cite{kossen2023context, chanDataDistributionalProperties2022}, i.e., the traditional gradient-based updates of the models' parameters using Stochastic Gradient Descent, the emergent ICL ability of LLMs relies on their strong prior knowledge for the provided instructions to perform the task in a zero-shot or a few-shot manner. Therefore, studying the interplay between ICL and in-weights learning is important for our understanding of the behavior of LLMs on downstream tasks.

\begin{figure}[!tp]
\centerline{\includegraphics[scale=0.4]{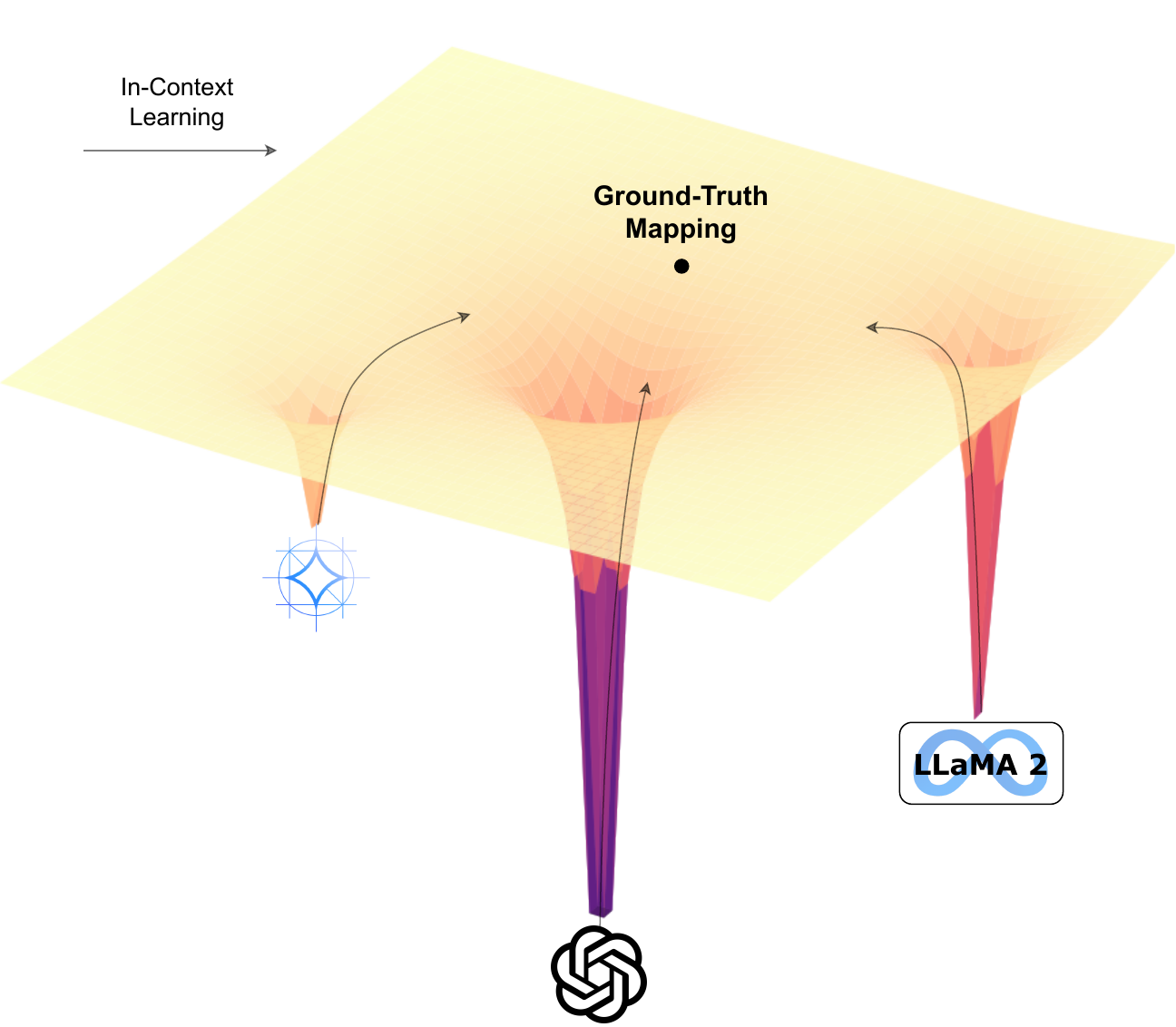}}
\caption{Knowledge priors in LLMs form ``gravity wells'' in the function space of the model. Larger models have deeper priors. Smaller LLMs have shallower but usually weaker priors. If the mapping for a specific task deviates from the existing knowledge prior, we find that In-Context Learning fails to nudge LLMs out of their knowledge priors enough to perform the task successfully.}
\label{fig:thumbnail}
\end{figure}

Prior work found evidence that LLMs may overly rely on their prior knowledge for the tasks to the detriment of their ICL demonstrations. Specifically, \cite{min2022rethinking} demonstrated that various LLMs, across different parameter scales, disregard the provided mapping from input to output in their instructions, and instead perform ``Task Recognition''; they focus on the source domain of the examples and the labels \textit{separately} to fetch the underlying task \cite{xie2021explanation}. Their experimental design consisted of randomizing the labels in the prompt, with minimal impact on performance. Since no ground truth is required, the authors propose this setting as an alternative ``zero-shot'' inference technique, the \textbf{task-recognition zero-shot}. Follow-up work has further studied this phenomenon by manipulating variants of the aforementioned experimental design, such as by extending the number of examples provided, using consistent permutations of the labels, using semantically unrelated substitute labels, etc.~\cite{kossen2023context, wei2023larger, pan2023context}, partly challenging their findings.
While we find that the common underlying implication of the literature revolves around the \textit{strong pull of the priors} on the final predictions, there have been no attempts to quantify this pull, especially in more complex settings.

To address this gap, in this work, we pivot from  the toy setting of binary classification of sentiment to multilabel emotion recognition \cite{demszkyGoEmotionsDatasetFinegrained2020, mohammad2018semeval}. We also experiment with LLMs of different scales, including both open-source and API-based models. Interestingly, we find that all these models significantly underperform simple BERT-based alternatives \cite{chochlakisLeveragingLabelCorrelations2023}, and have minor differences between them, both findings warranting further study of the underlying causes.

Consequently, we hypothesize that for complex subjective tasks where input-output mappings can vary substantially in different settings, prediction performance will stagnate due to the model's prior knowledge, with the phenomenon becoming stronger as the model's scale increases.

To evaluate this hypothesis, we propose an experimental design and measurements to quantify the \textbf{pull} and \textbf{consistency} of prior knowledge in various LLMs. We define the \textbf{pull} of prior knowledge as the similarity of the outputs of an LLM in the ICL setting with ``zero-shot'' predictions. We contextualize this quantity by comparing it to the similarity of the predictions with the ground truth, i.e., the model's performance. We define the \textbf{consistency} of the prior in two ways. First, as the similarity of the ``zero-shot'' predictions of the model using different prompt configurations, compared to that of the regular ICL predictions of the model. This evaluates whether we can extract true priors in the Bayesian sense of the term, i.e., not influenced by the particular instantiation of the prompt given a particular task. In the second definition of consistency, we examine our ability to create a feeback loop by prompting the model with its own predictions. In this manner, we examine the upper bounds of the model's performance assuming that its existing prior knowledge would be the easiest to follow and predict out of any other possible valid mapping.

We show that LLMs have \textit{inconsistent} priors even with 0 temperature, with similarity varying significantly between ``zero-shot'' runs, on par with the consistency of the regular few-shot ICL. Nevertheless, LLMs have strong enough task priors that their outputs reliably resemble the task-recognition zero-shot more than the ground-truth labels provided for ICL. Additionally, we observe that larger language models suffer more from a pull from the task priors, contrary to previous results indicating better ICL abilities in simpler settings. Overall, \textit{our experiments fail to reject our hypothesis}; in contrast, they provide further evidence in support of it.

Our novel experimental design and results contribute to the study of the trade-off between in-weights learning and in-context learning (or task recognition and task learning) in LLMs. Our results highlight the need for alternative ways to quantify bias in LLMs, and urge caution to researchers using ICL with LLMs to perform emotion recognition or other subjective tasks, where the input-output mappings are complex and can radically deviate from the model's pretraining distribution. Finally, they emphasize the need for and provide guidance into how to finetune LLMs to be less rigid and more adaptable to new information for subjective tasks.

\section{Related Work}

\subsection{In-Context Learning}

Since the emergence of the ICL technique \cite{brownLanguageModelsAre2020}, it has been widely used for LLM evaluations on public benchmarks~\cite{brownLanguageModelsAre2020, srivastava2022beyond} in few-shot or zero-shot settings, but has also been applied to derive insights from big data in other disciplines, such as computational social sciences \cite{ziems2023can} and health \cite{ kim2024health}. When used in combination with LLMs, it requires no finetuning, which is usually costly to perform for large models, and achieves competitive or state-of-the-art performance. Combined with the existence of  APIs~\cite{achiam2023gpt} or open-source implementations and models \cite{touvronLlamaOpenFoundation2023, Touvron2023a, zheng2023judging}, ICL presents an accessible and scalable \textit{passe-partout} for natural language processing tasks.

Researchers have studied many aspects of ICL. For example, \cite{chanDataDistributionalProperties2022} construct artificial scenarios and contrast ICL with in-weights learning by controlling the distribution of the training data, \cite{liu2022makes, rubin2022learning} examine how to best select demonstrations for the prompt, \cite{touvronLlamaOpenFoundation2023, zheng2023judging} integrate instructions explicitly during training, \cite{weiChainThoughtPrompting2022, yao2024tree} integrate reasoning to improve the robustness and the accuracy of the predictions of the model, \cite{zhao2021calibrate} examine ``cognitive'' biases that arise in LLMs, \cite{xie2021explanation} propose a Bayesian inference framework for interpreting ICL, etc.

Relatedly, researchers have also looked at the priors of LLMs. \cite{min2022rethinking} sampled random labels for the examples of the prompt, and found that the model's performance degraded only slightly, or even increased in some settings. This result suggested that LLMs recognize the task in the prompt more so than learn from it, and thereafter perform inference using their prior knowledge of the task (``Task Recognition'' rather than ``Task Learning''). Since no annotations are required for such a setting, and the performance is improved compared to zero-shot inference, the authors suggest that this inference mode can serve as a better, less naive ``zero-shot'' baseline.

Subsequent results \cite{kossen2023context} challenged the view that LLMs mostly perform task recognition, showcasing a significant degradation in performance when increasing the number of randomized examples in the prompt. To further disentangle task recognition and task learning, researchers have also used substitute words in place of the original labels, like \texttt{A} and \texttt{B} \cite{kossen2023context, pan2023context}, or semantically unrelated words \cite{wei2023larger} in place of \texttt{positive} and \texttt{negative} for sentiment classification, or even flipped the labels in said binary classification task.

The explicit consensus of the literature is that task learning complements task recognition. Nonetheless, the fact remains that the models can perform the task with randomized labels until a certain threshold of examples. This indicates an ability to override the specified mapping, or rather an inability to follow instructions conflicting with the background knowledge for the task. Moreover, these phenomena have been studied in the toy setting of sentiment classification, a binary classification task. Finally, the literature has not quantified the extent to which background knowledge influences predictions in the regular ICL setting.

\subsection{Survey-based Prompting of Priors}

Previous work has examined the priors, predispositions, or biases of LLMs by prompting them with surveys in the multiple-choice format.  \cite{santurkarWhoseOpinionsLanguage2023a} prompt LLMs with Pew Research's American Trends Panels, and compare the responses of the models with those of the US population, focusing on which groups the models align with. \cite{yongsatianchot2023investigating} use similar techniques to prompt the emotional priors of the models. Most prominently, however, these techniques are used to quantify political bias in LLMs \cite{hartmann2023political}.
Similar to the spirit of ICL, researchers have proposed \textit{steering} language models to generate responses from the perspective of certain groups, specified by their demographic characteristics~\cite{zhao2023group}. However, other works have shown that steering does not work well to change responses to fit different groups~\cite{he2024whose}.

In contrast, we take a different approach to quantifying alignment by looking at the similarities of the model's predictions in ``data-driven'' machine learning datasets, and essentially benchmark the ability of ICL to nudge the model away from its prior.

\section{Methodology}

We introduce common notation to simplify the introduction of our methodology in this section. For a set of examples $\mathcal{X}$, and a set of labels $\mathcal{Y}$, a dataset $\mathcal{D}$ defines a mapping $f:\mathcal{X} \rightarrow \mathcal{Y}$, and therefore \mbox{$\mathcal{D} = \{(x, y): x \in \mathcal{X}, y = f(x)\}$}, from which we can sample demonstrations with $p(x, y)$ (usually uniform over all text-label pairs). We do not differentiate between splits for brevity. Given prompt \mbox{$S = \{(x_i, y_i): (x_i, y_i) \sim p_S, i \in [k]\}$} with $k$ demonstrations from sampling distribution $p_S$, an LLM produces its own mapping and predictions for the task \mbox{$\hat{f}_k(.; p_S): \mathcal{X} \rightarrow \mathcal{Y}$}, $\hat{y} = \hat{f}_k(x; p_S)$. For all our experiments, we set the temperature to 0 to derive deterministic predictions. Following this notation, the regular ICL setting with $k$ demonstrations sampled from $\mathcal{D}$ with $p(x, y)$ is $\hat{f}_{k}(.; p)$.

\subsection{Similarity Measures}

\begin{figure}[!tp]
\centerline{\includegraphics[scale=0.63]{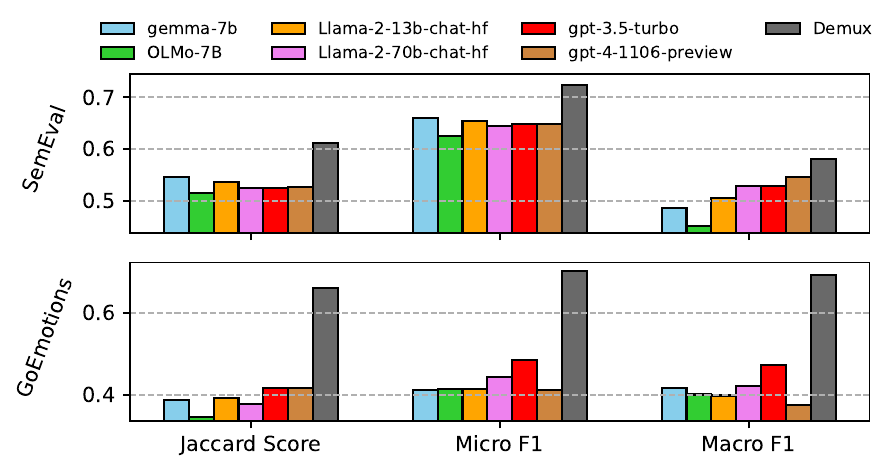}}
\caption{Comparison of performance of various LLMs with similarity-based retrieval for In-Context Learning and BERT-based Demux~\cite{chochlakisLeveragingLabelCorrelations2023} on SemEval 2018 Task 1 E-c~\cite{mohammad2018semeval} and GoEmotions~\cite{demszkyGoEmotionsDatasetFinegrained2020}.}
\label{fig:baselines}
\end{figure}

To evaluate closed-source, API-based models alongside open-source models, we rely on similarity measures calculated directly on the final predicted outputs, rather than using probabilistic measures like the models' output logits. 

Using probabilistic measures is not straightforward for these multilabel tasks. In single-label classification, one can take the maximum logit across labels, and normalize the logits to sum to 1. But there is no obvious normalization procedure for multilabel settings. Additionally, these models tend to output labels in the same order as provided in the prompt, introducing artifacts in the logit outputs tied to that input order.

Therefore, we use the Jaccard Score, Micro F1, and Macro F1 metrics~\cite{mohammad2018semeval} to evaluate the performance of the models. For consistency, we also use them as proxies to quantify the similarity between the prediction sets from different model runs. Crucially, they are symmetric functions, allowing us to apply them to interchangeable prediction sets.

\subsection{Task Prior Knowledge Proxies via Zero-Shot Inference}

We use various ``zero-shot'' inference setups in an attempt to elicit the task priors of each model. The most straightforward one is the \textit{regular zero-shot} inference, where the prompt contains instructions followed by the query only, yielding the mapping $\hat{f}_{0}(x; p_S) = \hat{f}_0(x)$ and \mbox{$\mathcal{D}_0 = \{(x, y): x \in \mathcal{X}, y = \hat{f}_0(x)\}$}.
Additionally, we use the \textbf{task-recognition zero-shot}\footnote{we assume that using unlabeled training data is permitted}, where the prompt contains $k$ demonstrations sampled with \mbox{$p_T(x, y) = p(x) q(y)$}. Labels are chosen independently from text, and hence they are random. We use two different variations of $q$: first, we simply set $p_T^I(x, y) = p(x) p(y)$, hence sampling results in \textit{independent} sampling of text and labels from $\mathcal{D}$. This effectively maintains the higher-order relationships between labels, which are strong in multilabel emotion recognition. We also experiment with a $q$ that is label-agnostic, where we sample each category \textit{uniformly} with independent 50\% binomial trials, which we denote as $p_T^U(x, y)$. We can construct $\mathcal{D}_k^I = \{(x, y): x \in \mathcal{X}, y = \hat{f}_{k}(x; p_T^I) \}$ and $\mathcal{D}_k^U = \{(x, y): x \in \mathcal{X}, y = \hat{f}_{k}(x; p_T^U) \}$. We refer to $\hat{f}_{k}(.; p_T^I), \hat{f}_{k}(.; p_T^U),$ and $\hat{f}_{0}$ as \textbf{task priors}.


\subsection{Improvement over Task Priors}

To establish whether ICL provides any gains in performance over the proxy task priors of each model, we compute the percentage improvement of ICL ($\hat{f}_{k}(.; p)$) over the best performing task priors over all examined $k$.

\subsection{Pull of Task Priors} \label{sec:pull}

\begin{figure*}[htbp] 
\centerline{\includegraphics[scale=0.6]{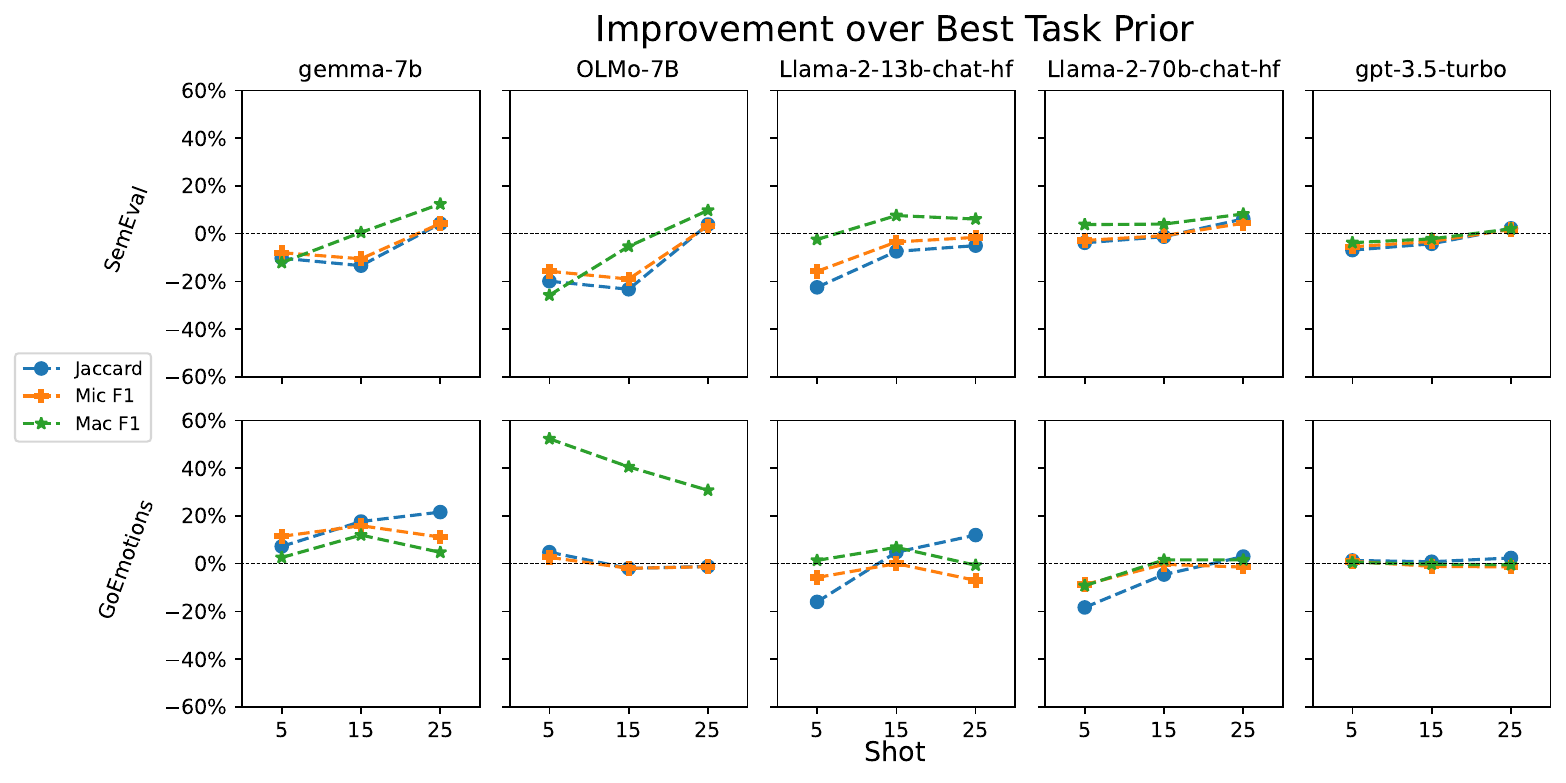}}
\caption{Percentage improvement of ICL over the performance of the corresponding model's best zero-shot inference mode (here: task-recognition zero-shot inference with $p_T^I$) in terms of Jaccard Score~(\texttt{Jaccard}), Micro F1~(\texttt{Mic F1}) and Macro F1~(\texttt{Mac F1}).}
\label{fig:improvement-over-prior}
\end{figure*}

To quantify the pull of the prior, we compare the similarity of $\hat{f}_k(.; p)$ with the ground truth (that is, the ICL performance of the model) to the similarity of $\hat{f}_k(.; p)$ with the task priors. We use the following sampling strategy, denoted as \texttt{Prior}; for each run through the evaluation set, we resample examples using $p(x)$ and labels with $q(y)$. Higher similarity with the task prior indicate a greater pull of knowledge priors on the final predictions. To compare across models, we can use the difference between the two similarities for each.

\subsection{Consistency} \label{sec:consistency}

\begin{figure}[!b] 
\centerline{\includegraphics[scale=0.66]{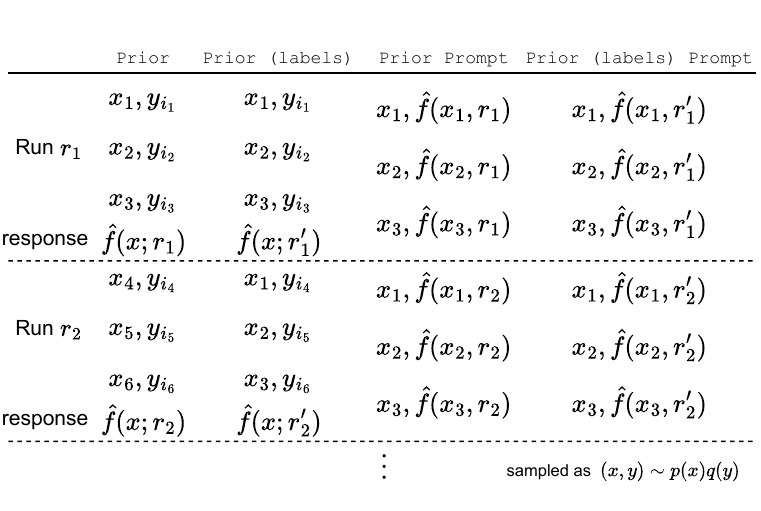}}
\caption{Qualitative demonstration of our text-label sampling schemes. Shown are the prompt ``demonstrations'' for all queries of the evaluation set per run.}
\label{fig:sampling}
\end{figure}

In quantifying the consistency of the priors, multiple runs are required, in which our main focus is to vary some aspect of the prompt to check the similarity of the predictions across them. First, we perform this analysis on the regular ICL setting $\hat{f}_{k}(.; p)$ by resampling from $p$. Then, for the proxy task priors, we check consistency in two sampling settings from $p_T$. One is \texttt{Prior} from Section \ref{sec:pull}. In the second, \texttt{Prior (labels)}, we sample the same examples for all runs from $p(x)$, and then, for each run, we sample labels for them from $q(y)$. In this manner, we evaluate how the examples and the labels separately affect the task-recognition zero-shot inference. While different examples in \texttt{Prior} could provide different evidence for the domain of the problem (e.g., the presence of special characters like hashtags for social media text), a true task prior should be minimally impacted by different random labels, as examined in \texttt{Prior (labels)}.
We cannot perform this analysis on the regular zero-shot $\hat{f}_0$.

We also examine the ICL setting where the labels in the prompt have been replaced by the task prior predictions of the model. In other words, we reinforce the model's prior in its prompt, and check whether that breeds consistency. Concretely, in the ICL setting $\hat{f}_{k}(.; p)$, instead of sampling text-label pairs from $\mathcal{D}$, we can sample from either $\mathcal{D}_0, \mathcal{D}_k^I,$ or $\mathcal{D}_k^U$. We again construct two settings, \texttt{Prior Prompt} and \texttt{Prior (labels) Prompt}. Each samples from a dataset constructed by the corresponding method described above, \texttt{Prior} and \texttt{Prior (labels)} respectively. For both, our sampling strategy involves sampling examples once, and then selecting their predictions from the various different runs of the corresponding prediction method. An illustration of the sampling for all described methods can be seen in Fig. \ref{fig:sampling}.

\subsection{Proxy performance}

\begin{figure*}[htbp]
\centerline{\includegraphics[scale=0.6]{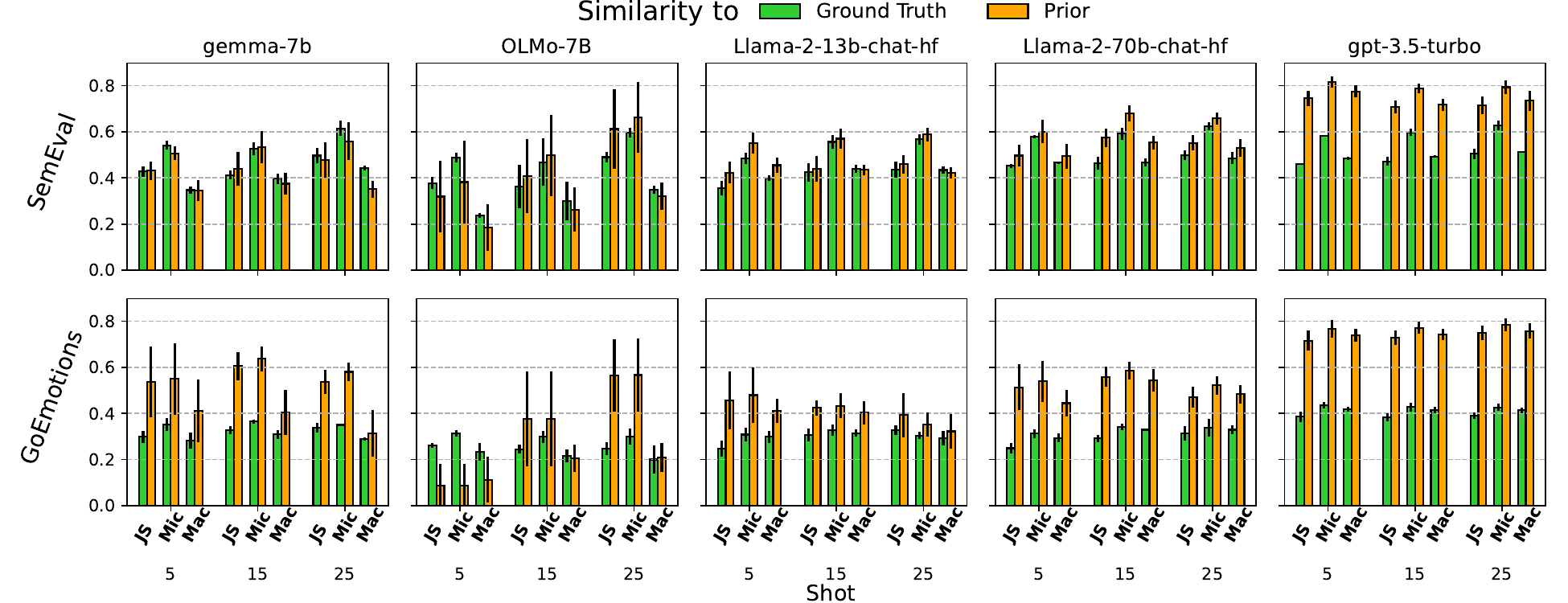}}
\caption{Comparison of similarity of model's predictions with the ground truth and with the task-recognition zero-shot predictions of the same shot, quantifying our notion of \textbf{pull of prior knowledge} in terms of Jaccard Score (\texttt{JS}), Micro F1 (\texttt{Mic}) and Macro F1 (\texttt{Mac}).}
\label{fig:pull}
\end{figure*}

For the latter two settings, we can also look at the performance of the model with respect to the dataset used to sample demonstrations from instead of the ground truth, and compare it to the performance of the regular ICL model $\hat{f}_{k}(.;p)$. This can be thought of as a performance upper bound for a valid mapping, assuming that the model's task priors would be the easiest to predict out of any other possible valid mapping.

\subsection{Prompt Design}

Previous work has demonstrated that small changes in the prompt can have a significant impact on the outputs of LLMs, such as recency and other biases~\cite{zhao2021calibrate}, the prompt template (as we also find), or even the inclusion of a seemingly random sequence of tokens~\cite{zou2023universal}. In our effort to reduce the search space, we standardize the prompt template, presenting the one that yields the best performance with respect to the ground truth among the ones we experimented with. In addition, because the specific examples and their order in the prompt can affect the output of the model, we use exactly the same examples, in the same order across corresponding experiments.

\section{Experiments} \label{sec:exps}

\subsection{Datasets}

We use two multilabel emotion recognition, \mbox{SemEval 2018 Task 1 E-c}~\cite{mohammad2018semeval} and GoEmotions~\cite{demszkyGoEmotionsDatasetFinegrained2020}, both of which contain social media posts.

\noindent \textbf{SemEval 2018 Task 1 E-c}: Multilabel emotion recognition benchmark containing annotations for 11 emotions: \textit{anger}, \textit{anticipation}, \textit{disgust}, \textit{fear}, \textit{joy}, \textit{love}, \textit{optimism}, \textit{pessimism}, \textit{sadness}, \textit{surprise}, and \textit{trust}. The dataset is comprised of tweets in three languages, but we use only the English subset. To reduce our inference costs, we evaluate the models on the smallest split, which is the development set of the dataset, containing 886 examples. We refer to this as \textbf{SemEval}.

\noindent \textbf{GoEmotions}: Multilabel emotion recognition benchmark containing annotations for the 27 emotions proposed by \cite{cowenSelfreportCaptures272017}. The dataset contains Reddit comments in English. To reduce our inference costs, and for consistency, we evaluate the model on a random subset of 800 examples from the development set. To further reduce costs, but also make the task easier for the LLMs, we pool the emotions to the following 7 ``clusters'' by using the hierarchical clustering provided by \cite{demszkyGoEmotionsDatasetFinegrained2020}: \textit{admiration}, \textit{anger}, \textit{fear}, \textit{joy}, \textit{optimism}, \textit{sadness}, and \textit{surprise}.

\subsection{Implementation Details}

\begin{figure*}[htbp]
\centerline{\includegraphics[scale=0.5]{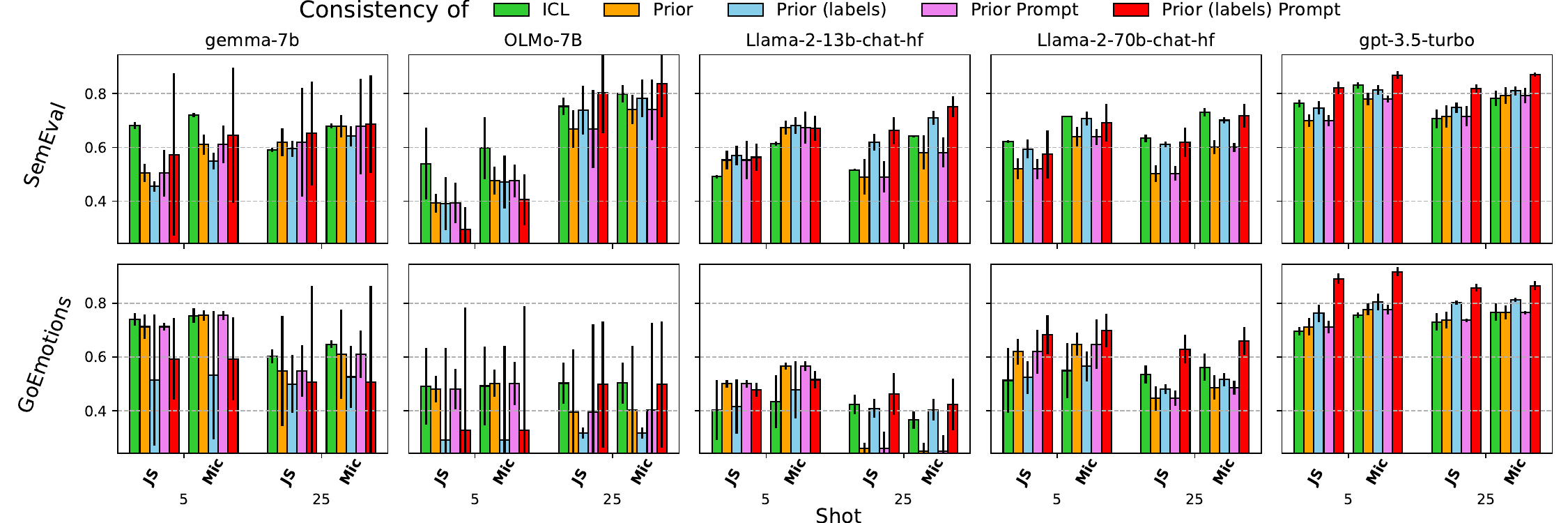}}
\caption{Comparison of consistency between different runs for the regular \texttt{ICL} setting, the task prior $f_k(.; p_T^I)$ with different examples in the prompt (\texttt{Prior}) or the same examples but different random labels (\texttt{Prior (labels)}), and the ICL setting where the labels have been replaced with the task-recognition zero-shot inference predictions with $p_T^I$ from \texttt{Prior}, denoted \texttt{Prior Prompt}, or \texttt{Prior (labels)}, denoted \texttt{Prior (labels) Prompt}. Quantified as the similarity between the predictions of the models across runs in terms of Jaccard Score (\texttt{JS}) and Micro F1 (\texttt{Mic}). 15-shot and Macro F1 skipped to avoid clutter.}
\label{fig:consistency}
\end{figure*}

We set the temperature of the LLMs to 0. We use the 4-bit quantized versions of the open-source LLMs through the \textit{HuggingFace}~\cite{wolf-etal-2020-transformers} interface for \textit{PyTorch}. We use Gemma (\texttt{google/gemma-7b}), OLMo (\texttt{allenai/OLMo-7B}), LLaMA 2 13b (\texttt{meta-llama/Llama-2-13b-chat-hf}) and 70b (\texttt{meta-llama/Llama-2-70b-chat-hf}). For API-based models, we use GPT-3.5 (\texttt{gpt-3.5-turbo}), and we do some baselining with GPT-4 (\texttt{gpt-4-1106-preview}). We use the open-source implementation provided by \cite{chochlakisLeveragingLabelCorrelations2023} for Demux. We perform 3 runs for each LLM experiment, varying the examples used or the labels for them, as described in Section \ref{sec:consistency}. We control which examples or labels are selected for each run to ensure consistency across models. We present mean and standard deviation where available. Comprehensive results are presented in the Appendix. When computing means and standard deviations of similarities, we use every possible pair between two configurations. Demux results are averages across multiple runs for the same evaluation sets. We use random retrieval of examples, unless stated otherwise. We perform similarity-based retrieval \cite{liu2022makes} with \texttt{sentence-transformers/all-mpnet-base-v2}~\cite{song2020mpnet}.



We use the following prompt template: \textit{ \footnotesize Perform emotion classification in the following examples by selecting none, one, or multiple of the following emotions: \{labels\}\textbackslash n\textbackslash nInput: \{text\}\textbackslash n\{label\}}. We found chat templates perform worse in the LLaMAs, whereas the GPTs showed no significant differences. For open-source models, we specifically formatted the labels as a JSON object to improve their ability to follow instructions. The GPTs showed no such issues, hence we used a comma-separated format.

\subsection{Baselining}

First, we present the best performance of each model for each benchmark in terms of Jaccard Score, Micro F1 and Macro F1, and compare that to Demux~\cite{chochlakisLeveragingLabelCorrelations2023}, a BERT-based alternative. To elicit the best performance from the LLMs, we use similarity-based retrieval. Results are shown in Fig. \ref{fig:baselines} for 25-shot LLMs. We validate that performance does not scale significantly with more demonstrations in the Appendix.

We observe that LLMs underperform the BERT-based alternative significantly, especially for GoEmotions, where Demux improves by more than 50\%. We also note that the scale of the models does not seem to be a major factor in their final performance. In SemEval, for example, we find that the two LLaMA 2 models and the two GPT models perform equivalently, and that Gemma performs on par with them.

We conclude that for emotion recognition, ICL seems to significantly lag behind traditional gradient-based alternatives. The evidence thus far suggests that the prior knowledge of LLMs on these tasks is interfering with their ability to perform them. We present further evidence that this is indeed the case.

\subsection{Differences between Task Priors}

Results for $\hat{f}_0$ and $\hat{f}_k(.; p_T^U)$ are weaker than $\hat{f}_k(.; p_T^I)$, both in terms of predicting the ground-truth labels, but also with respect to the similarity to other predictions, as can be seen in the Appendix. Therefore, we present results only for the more grounded and realistic prior, $\hat{f}_k(.; p_T^I)$.

\subsection{Improvement over Task Priors} \label{sec:improve-prior}

\begin{figure*}[htbp]
\centerline{\includegraphics[scale=0.48]{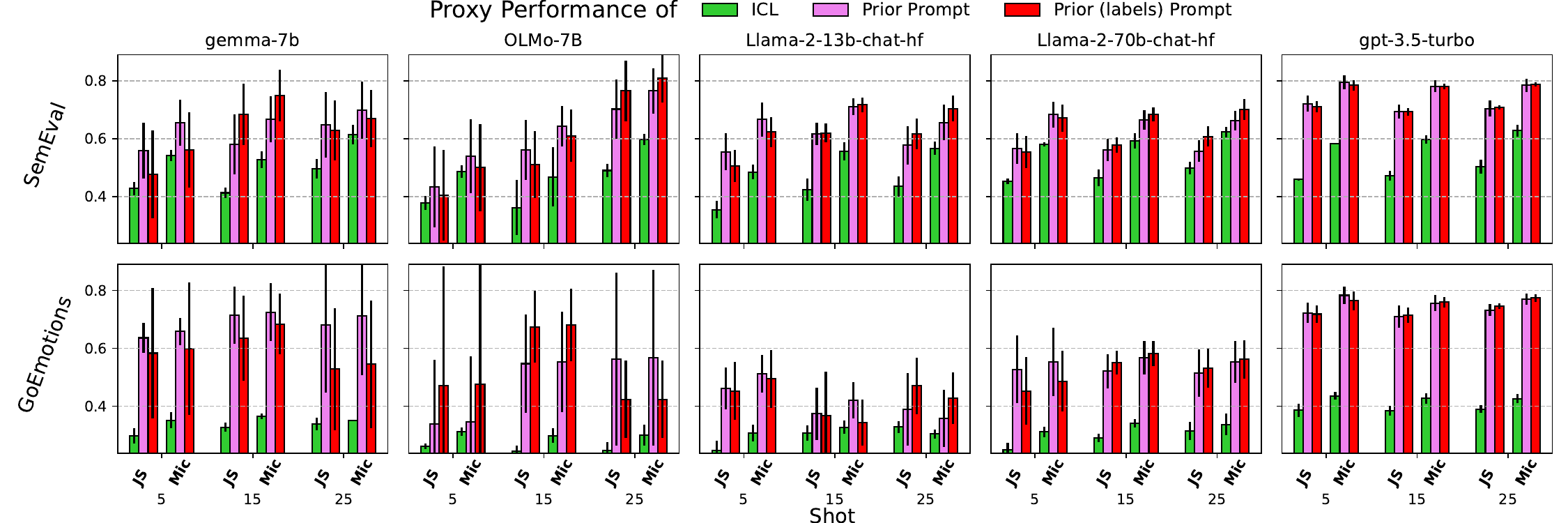}}
\caption{Comparison of proxy performance in terms of Jaccard Score (\texttt{JS}), Micro F1 (\texttt{Mic}) and Macro F1 (\texttt{Mac}). We compare the regular \texttt{ICL} setting, \texttt{Prior Prompt} which samples demonstrations from $\mathcal{D}_k^I$ created by \texttt{Prior}, and \texttt{Prior (labels) Prompt} which samples demonstrations from $\mathcal{D}_k^I$ created by \texttt{Prior (labels)}. The performance is computed against the datasets used to sample the labels from, which means that models prompted with the task prior are also evaluated against the task prior predictions. Macro F1 skipped to avoid clutter.}
\label{fig:reinforcement-accuracy}
\end{figure*}

First, we examine what the improvements in performance with ICL are compared to the performance of the priors. In Fig. \ref{fig:improvement-over-prior}, we present the percentage improvement as we vary the number of examples in the prompt compared to the best-performing task prior within that range. We observe that there are few instances where the models actually improve over zero-shot inference. Moreover, there is a marked trend of grounding to 0 instead of improving as the scale of the models' increases. For example, for both benchmarks, we see that \mbox{GPT-3.5} has essentially equivalent performance to the prior for all configurations, while LLaMA 2 70b also flattens out compared to its smaller, 13b counterpart.

The largest supervised models, therefore, seem to be achieving ``random'' levels of performance, indicating that for the task of multilabel emotion recognition, ICL may be providing little to no benefit above the prior knowledge of the model. We present additional experiments bolstering that picture.

\subsection{Pull of Task Priors} \label{sec:pull-exps}

We examine this phenomenon by looking at the pull of the priors, as we had defined in Section \ref{sec:pull}. We present our findings in Fig.~\ref{fig:pull}, where we contrast the similarity of the predictions to the ground truth (equivalent to the performance) of the model to the similarity of the predictions to the task prior $\hat{f}_k(.; p_T^I)$. We observe that the models tend to have a positive prior pull (meaning higher similarity to the proxy task prior), with the size of the models increasing its pull. Note that this occurs while we provide the ground truth in the prompt, not the predictions (Sections \ref{sec:consistency-exps} and \ref{sec:reinforcement-exps}).

In particular, we see that GPT-3.5 displays a very large preference towards its prior knowledge, explaining the grounding of its improvement to 0. Similarly, LLaMA 2 70b shows a larger prior pull compared to LLaMA 2 13b. Finally, we see that only smaller models can achieve negative prior pull. We also see that there is no general trend towards decreasing similarity to the prior as the number of demonstrations increases. The same trends are shown for higher shots in the Appendix.

\subsection{Consistency} \label{sec:consistency-exps}

We now look at the consistency of the task priors, defined in Section \ref{sec:consistency}. We present results when sampling from $\mathcal{D}_k^I$. Results for $\mathcal{D}_0$ can be seen in the Appendix. Higher consistency for the task priors compared to the regular predictions could explain part of the trend in Sections \ref{sec:improve-prior} and \ref{sec:pull-exps}, since less erratic predictions could achieve higher average performance.

Results are shown in Fig. \ref{fig:consistency}. We see no evidence for higher task prior consistency. In fact, we observe many occasions where the consistency of the regular setting is indeed higher, yet the model's predictions are more similar to the task prior than the ground truth. Such is the case for GPT-3.5 and LLaMA 2 70b in SemEval, for instance.

We do see a small trend among the task priors for \texttt{Prior (labels)}, and by extension, \texttt{Prior (labels) Prompt}, to be slightly more consistent, which conforms with our expectations. Overall, however, these indicate the proxy task priors are not true Bayesian priors. Nevertheless, the proxy task priors become more consistent as models get larger, suggesting that they will asymptotically converge to the priors.

\subsection{Proxy Performance} \label{sec:reinforcement-exps}

Finally, we examine whether the models can actually predict their proxy task prior when prompted with them. This proxy performance is presented in Fig. \ref{fig:reinforcement-accuracy} compared to the regular ICL setting. We observe that the proxy performance of the reinforced priors is higher. Notably, the performance of the model almost doubles for GPT-3.5 compared to the ground truth performance. This demonstrates that, in principle, LLMs can perform competitively on emotion recognition tasks.

For the larger models in this reinforcement setting, however, there seems to be \textit{no improvement in performance from additional examples} in the prompt. Our interpretation is that the models are, expectedly, still performing at ``random'' levels of performance because their prior knowledge of the task coincides with the mapping we use to evaluate them, and no new knowledge is provided to them in the prompt.

\section{Conclusion}

In this work, we 
examine LLMs and the effects of their prior knowledge in challenging multilabel emotion recognition benchmarks. We find that they severely underperform traditional BERT-based models in such settings, and explore our \textit{hypothesis} that this occurs due to the inability of In-Context Learning (ICL) to nudge the model away from its prior knowledge, or, in other words, because of the strong pull exerted by the prior knowledge on the model's prediction.

We find that, contrary to prior work in simpler settings, as the size of LLMs increases, they tend to be less amenable to ICL, instead defaulting to their prior knowledge for the task. Indeed, we show that the model's predictions more closely resemble the proxy priors than the ground-truth labels, with the resemblance increasing as model size increases. We further demonstrate that this happens despite the sensitivity of the task priors to the design of the ``Task Recognition'' prompt. Finally, we overall see more robust results on larger models.

We conjecture that this occurs because for complex subjective tasks such as emotion recognition, the mapping from the inputs to outputs might change drastically for different datasets. If the resulting mapping is sufficiently different from the knowledge priors of the model, as we demonstrate is the case for the benchmarks we examined, then the model mostly ignores the mapping, irrespective of the number of examples, and defaults to its internal mapping. On the other hand, if the mapping matches the in-weights knowledge, such as when we provide its proxy prior predictions in the prompt, then the model again simply uses its---now valid---internal mapping for prediction. Consequently, our experiments demonstrate that little can be gained in the vanilla ICL setting for highly subjective tasks, and shed light on the underlying reasons for the observed phenomena.



\section*{Ethical Statement}

Focusing on traditional machine learning benchmarks may take away attention from quantifying bias in LLMs using other tools and techniques \cite{caliskan2017semantics, gonen2019lipstick, ferrara2023should}. These types of analysis should compliment each other. It's important to note that improving affective capabilities in LLMs entails risks in addition to benefits, since better catering the emotional responses to more contexts can allow more capable manipulation of users by LLMs, and not necessarily only when deployed by bad actors.

We also delineate some pertinent limitations that could limit the generalizability of our findings. First, the datasets we utilized to benchmark our models have been criticized as lacking the appropriate context for a model---or even humans---to make appropriate judgments about the emotions expressed in them \cite{yang2023context}. Techniques to evaluate the difficulty or the correctness of the labels in a dataset have been devised based on prediction consistency and confidence \cite{swayamdipta2020dataset}. Given the lack of context and the subjectivity of the semantic interpretation of the inputs, such techniques could be used to exclude specific examples from the prompts and perhaps improve performance. That being said, the semantic interpretation by human annotators is consistent enough across examples for Demux to achieve reliably and significantly better performance than all the LLMs. While a limitation, it is congruent with our findings. Removing ambiguous examples may also remove all the nuances that make these datasets and tasks interesting \cite{aroyo2015truth}.

Furthermore, while we present additional results with higher shots in the Appendix for some of the models, results presented in Section \ref{sec:exps} include experiments with $\{5, 15, 25\}$ examples in the prompt. We do this due to computational constraints since, as we point out, we cannot use tricks similar to \cite{kossen2023context} to get performance at different shots simultaneously, get a prediction with a single output token, or use probabilistic measures owning to the multilabel nature of the datasets. Nonetheless, results in the Appendix do suggest that the effects remain robust as the number of examples increases, though only for task priors of appropriate ``shots'' \cite{kossen2023context}.

We reiterate that we have standardized the prompt template. It's also worth noting that we present results where we vary the prompt within the same configuration only three times. We made this choice to enable us to run multiple models instead of experimenting with fewer models or fewer aspects of their prior knowledge, notably the size of the model, and, consequently, can mostly speak of trends observed in the data.

Finally, we do not perform any finetuning on the LLMs. We also do not control for the amount of reinforcement learning from human feedback (RLHF)~\cite{ouyangTrainingLanguageModels2022} (or similar) that has been performed, which would be impossible for most closed-source models. It is possible that the priors can be overcome with enough training data to finetune LLMs, although the techniques used to efficiently do so might not have enough impact due to their sparseness~\cite{hu2021lora}. In addition, RLHF might modulate the phenomena we study. Nevertheless, our setup utilizes the main mode of using LLMs by the research community, especially for API-based models.

\bibliographystyle{IEEEtran}
\bibliography{library}

\clearpage

\begin{appendices}

\section{Detailed results}

We present comprehensive results for all our experiments between Figs. \ref{fig:js-gpt-semeval} and \ref{fig:mic-llama70-goemotions}. We present mean and standard deviation. The names of the columns/rows denote:
\begin{itemize}
    \item $X$s: $\hat{f}_X(.; p)$, $X$-shot random example retrieval
    \item Cossim-$X$s: $X$-shot similarity-based retrieval
    \item Random-prior-$X$s: $\hat{f}_{X}(.; p_T^U)$ task-recognition zero-shot with binomial sampling.
    \item Proxy-prior-$X$s: \texttt{Prior}, $\hat{f}_{X}(.; p_T^I)$ task-recognition zero-shot with independent sampling from $\mathcal{D}$.
    \item Prior-$Y$s-prompt-$X$s: \texttt{Prior Prompt}, for each example $e_i, i\in[X]$ in the prompt, we use $\hat{f}_Y(e_i, p_T^I)$ as its label.
    \item 0s-prompt-$X$s: For each example $e_i, i\in[X]$ in the prompt, we use the zero-shot prediction of the model $\hat{f}_0(e_i)$ as its label, aka sampling from $\mathcal{D}_0$
    \item Suffix \texttt{sedl}: \textbf{S}ame \textbf{E}xamples \textbf{D}ifferent \textbf{L}abels, indicates \texttt{Prior (labels)} or \texttt{Prior (labels) Prompt}.
    \item Suffix \texttt{traindev}: Whether the experiment includes predictions for the train split, where we use the test set of the dataset to retrieve examples.
\end{itemize}

\section{Main Text References}

\subsection{Baseline Scaling}

In Figs. \ref{fig:js-gpt-semeval} and \ref{fig:mic-gpt-semeval}, we can see that GPT-3.5 stops improving in performance with similarity-based retrieval on SemEval after 55 examples in the prompt. Similarly, LLaMA 2 13b stops scaling after 25 examples (the performance presented in the main text) on SemEval, as shown in Figs. \ref{fig:js-llama13-semeval} and \ref{fig:mic-llama13-semeval}.

\subsection{Different Priors}

Zero-shot prompting and Bernoulli-based sampling results are presented in all figures with some exceptions. For example, we skipped the Bernoulli-based prior in LLaMA 2 70b due to computational constraints, and do not present the zero-shot prior in smaller models because of their inability to follow instructions without any demonstrations.

\subsection{Prior Pull Trends for Higher Shots}

In Figs. \ref{fig:js-gpt-semeval} and \ref{fig:mic-gpt-semeval}, we see that for GPT-3.5 on SemEval, the similarity to the prior remains higher even for the \mbox{``Cossim-85s''} configuration compared to the ground truth. In fact, the highest similarity to the ground truth is smaller than the smallest similarity to the priors. The same trends are not observed in the much smaller LLaMA 2 13b in Figs.  \ref{fig:js-llama13-semeval} and \ref{fig:mic-llama13-semeval}.

\include{semeval-results}

\include{goemotions-results}

\end{appendices}

\end{document}

%% file: semeval-results.tex
\begin{figure*}[htbp] 
\centerline{\includegraphics[scale=0.17]{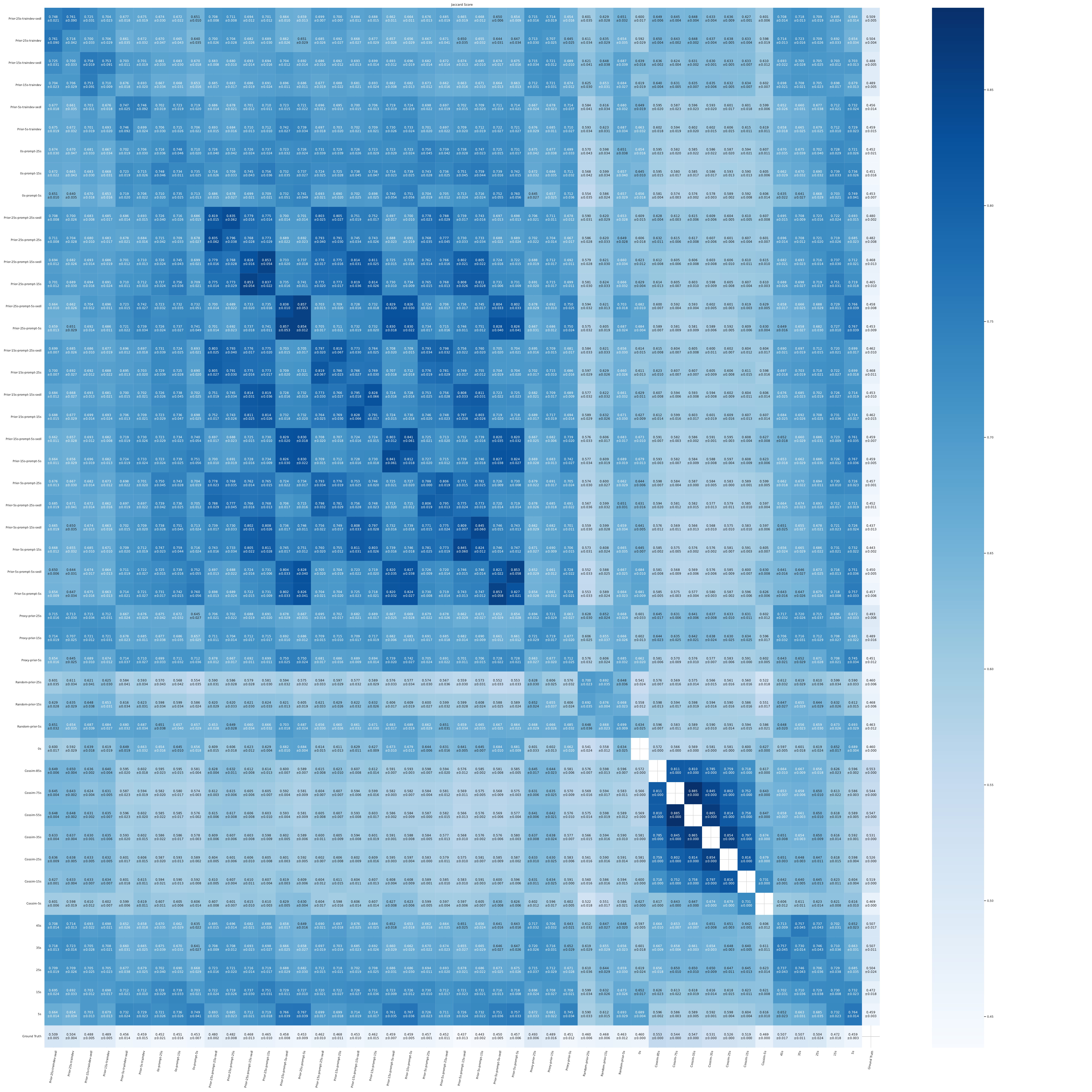}}
\caption{Jaccard Score similarities across \texttt{gpt-3.5-turbo} experiments on SemEval}
\label{fig:js-gpt-semeval}
\end{figure*}

\begin{figure*}[htbp] 
\centerline{\includegraphics[scale=0.17]{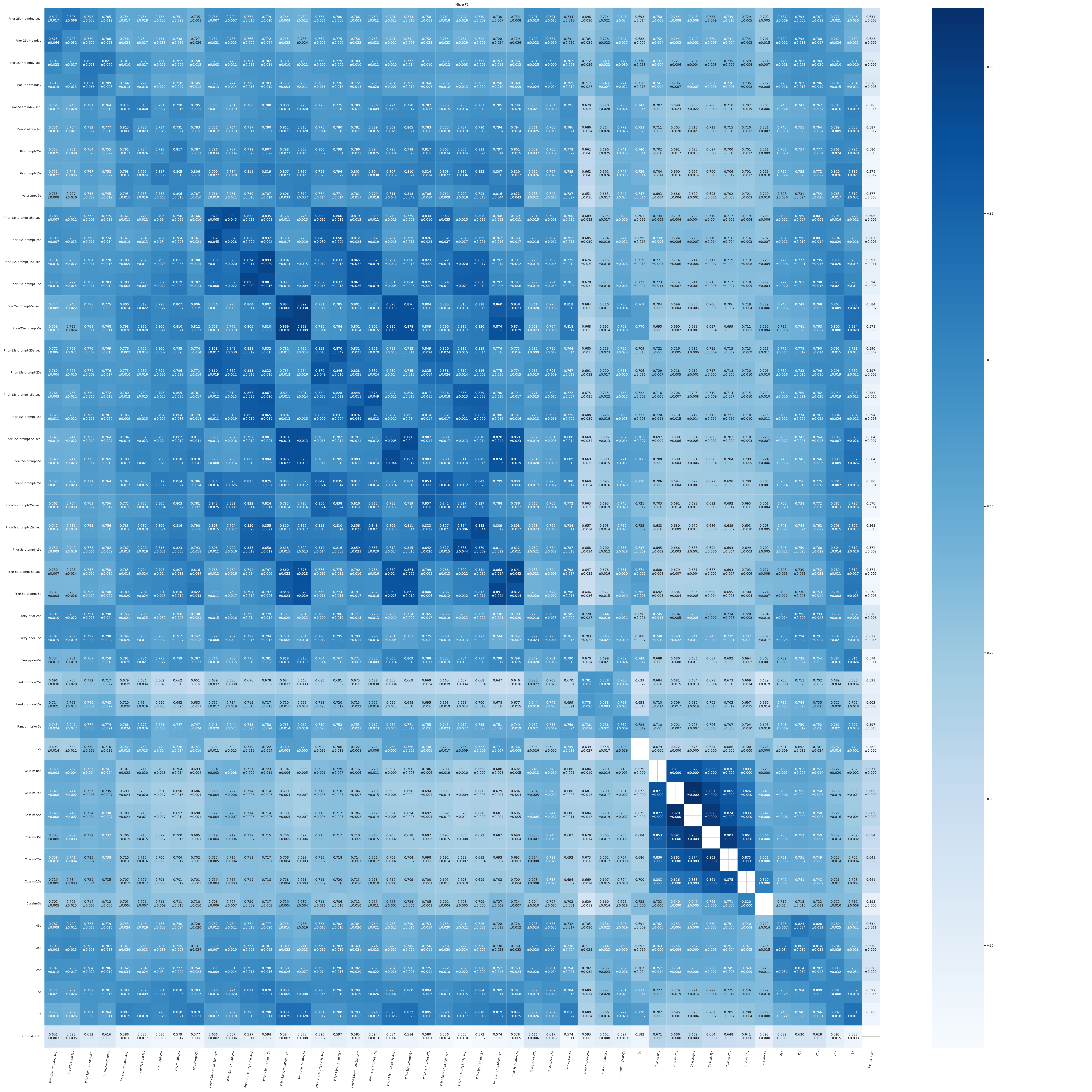}}
\caption{Micro F1 similarities across \texttt{gpt-3.5-turbo} experiments on SemEval}
\label{fig:mic-gpt-semeval}
\end{figure*}

\begin{figure*}[htbp] 
\centerline{\includegraphics[scale=0.2]{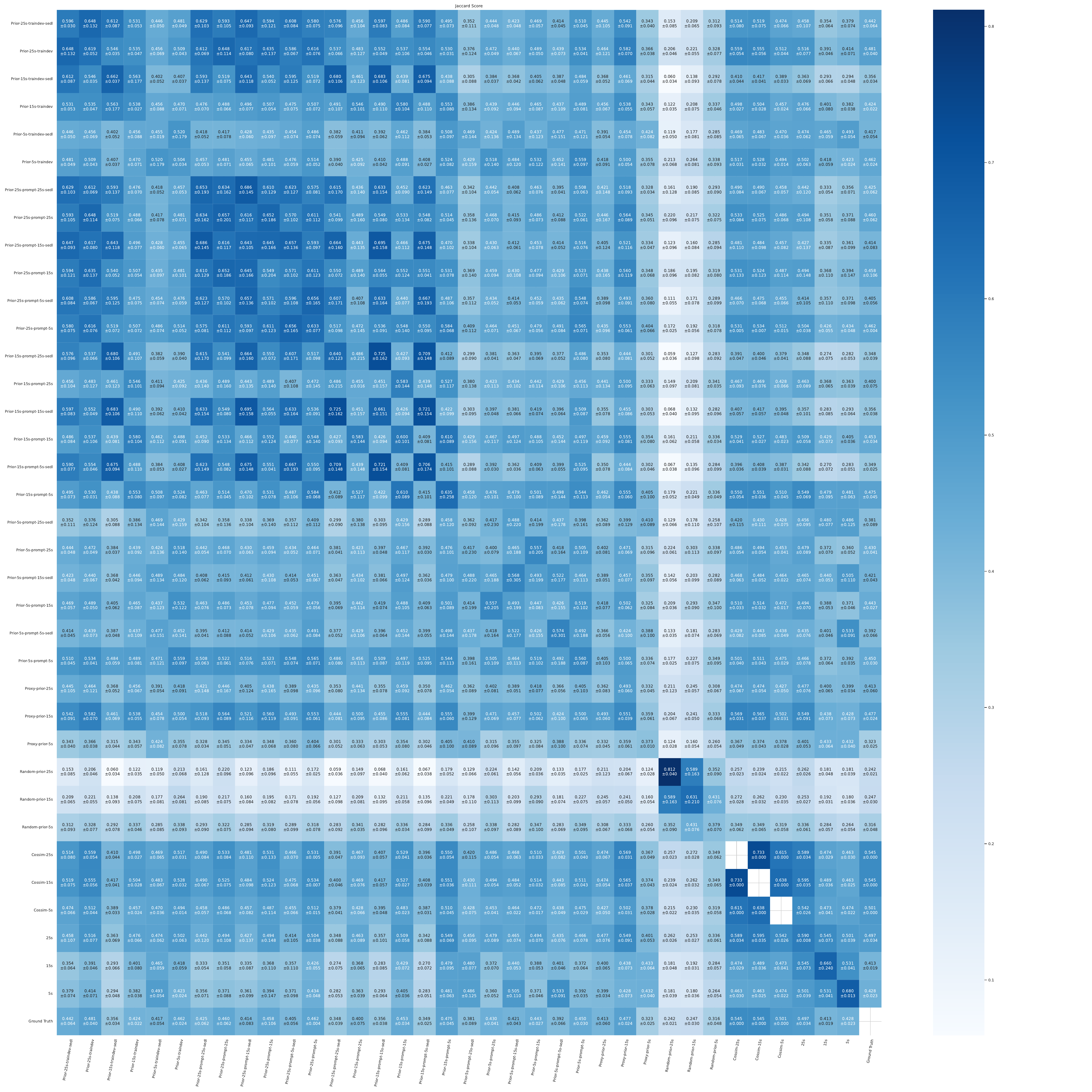}}
\caption{Jaccard Score similarities across \texttt{google/gemma-7b} experiments on SemEval}
\label{fig:js-gemma-semeval}
\end{figure*}

\begin{figure*}[htbp] 
\centerline{\includegraphics[scale=0.2]{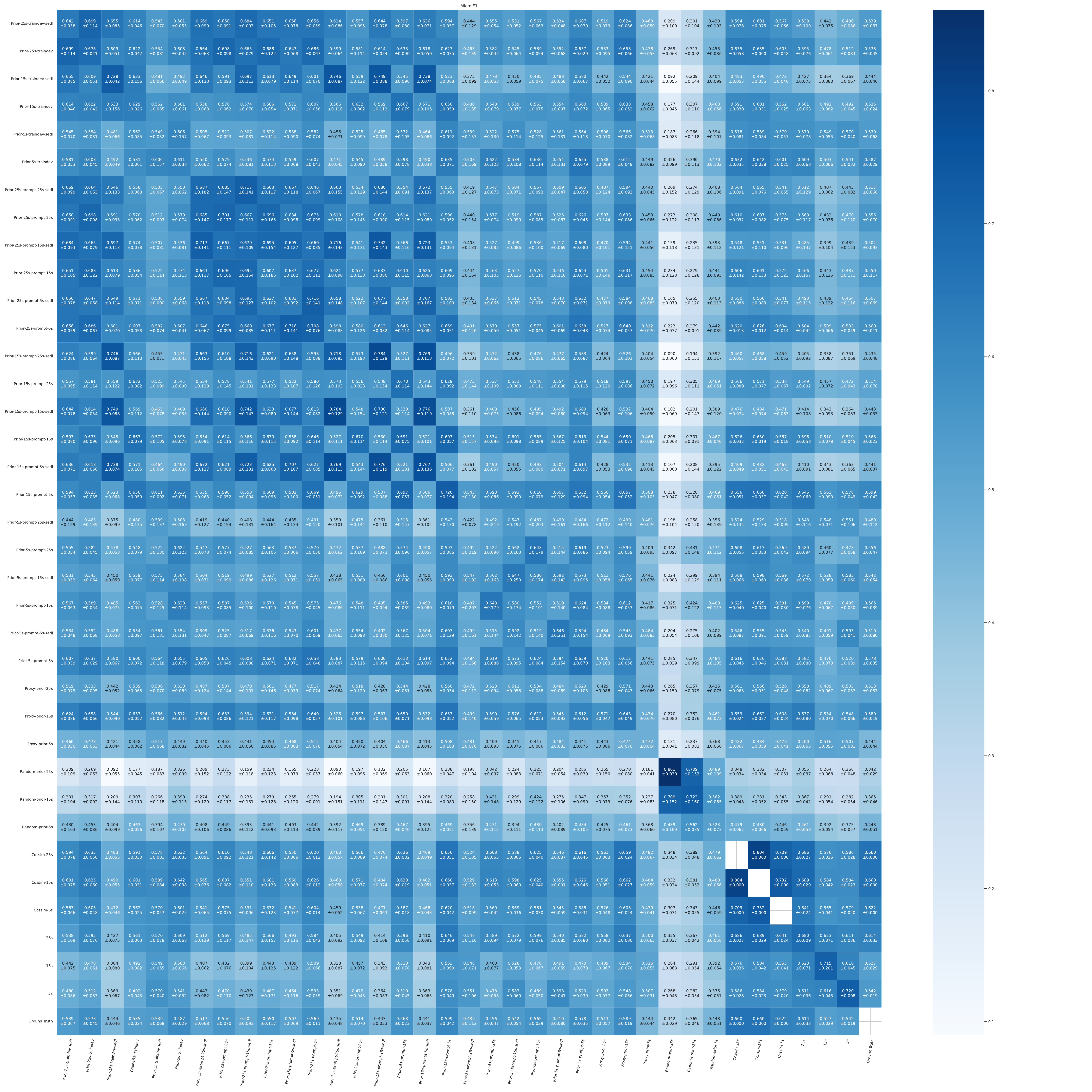}}
\caption{Micro F1 similarities across \texttt{google/gemma-7b} experiments on SemEval}
\label{fig:mic-gemma-semeval}
\end{figure*}

\begin{figure*}[htbp] 
\centerline{\includegraphics[scale=0.2]{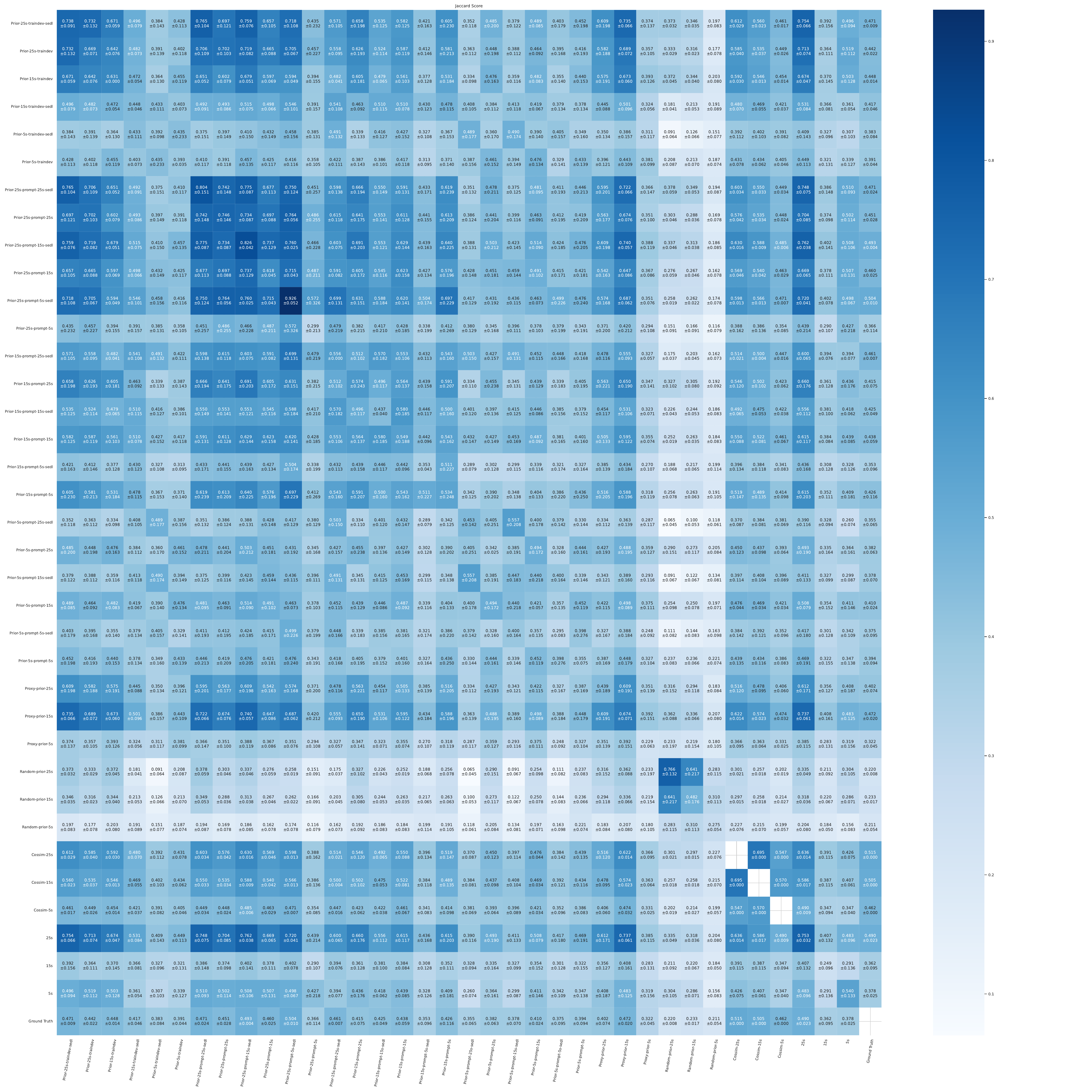}}
\caption{Jaccard Score similarities across \texttt{allenai/olmo-7B} experiments on SemEval}
\label{fig:js-olmo-semeval}
\end{figure*}

\begin{figure*}[htbp] 
\centerline{\includegraphics[scale=0.2]{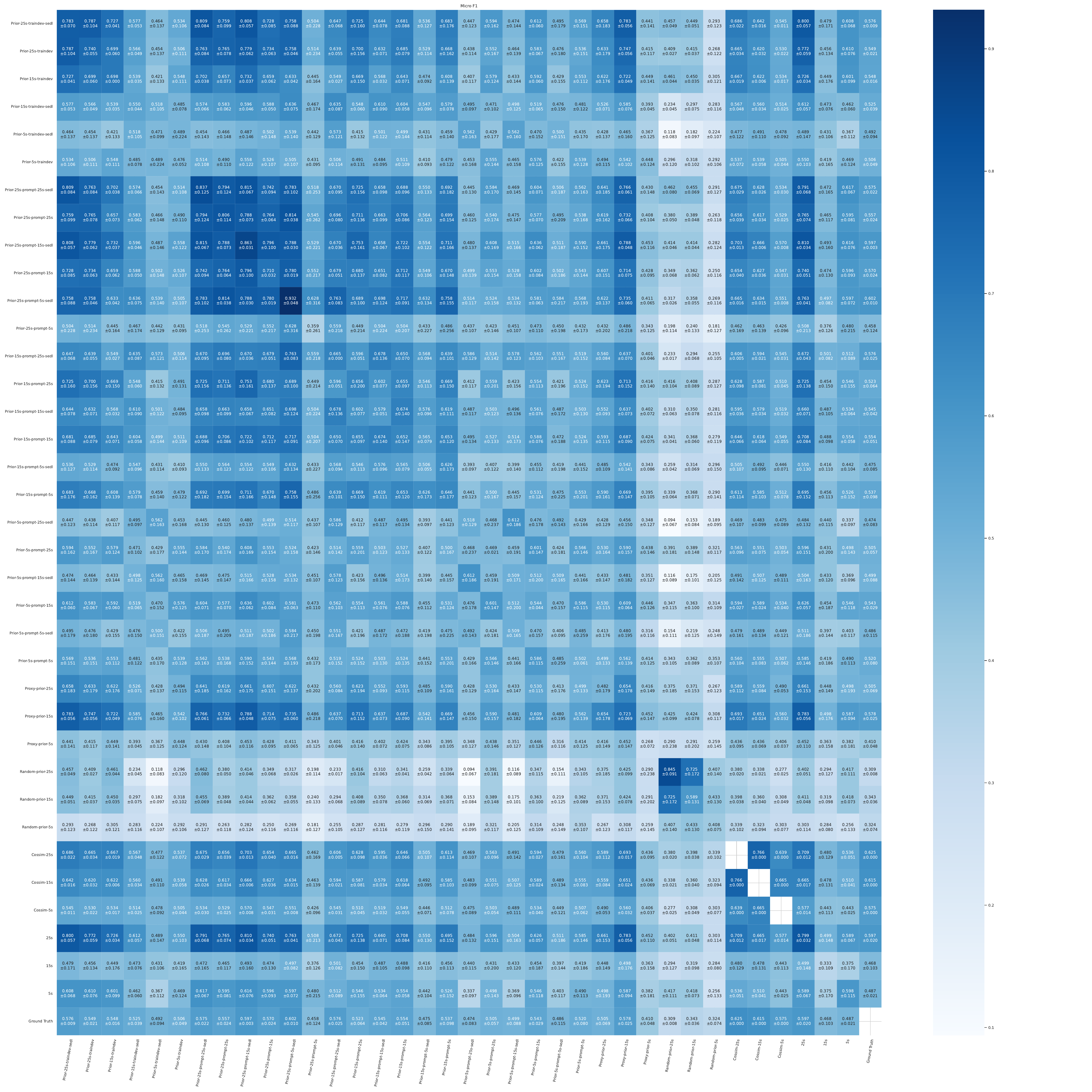}}
\caption{Micro F1 similarities across \texttt{allenai/olmo-7B} experiments on SemEval}
\label{fig:mic-olmo-semeval}
\end{figure*}

\begin{figure*}[htbp] 
\centerline{\includegraphics[scale=0.18]{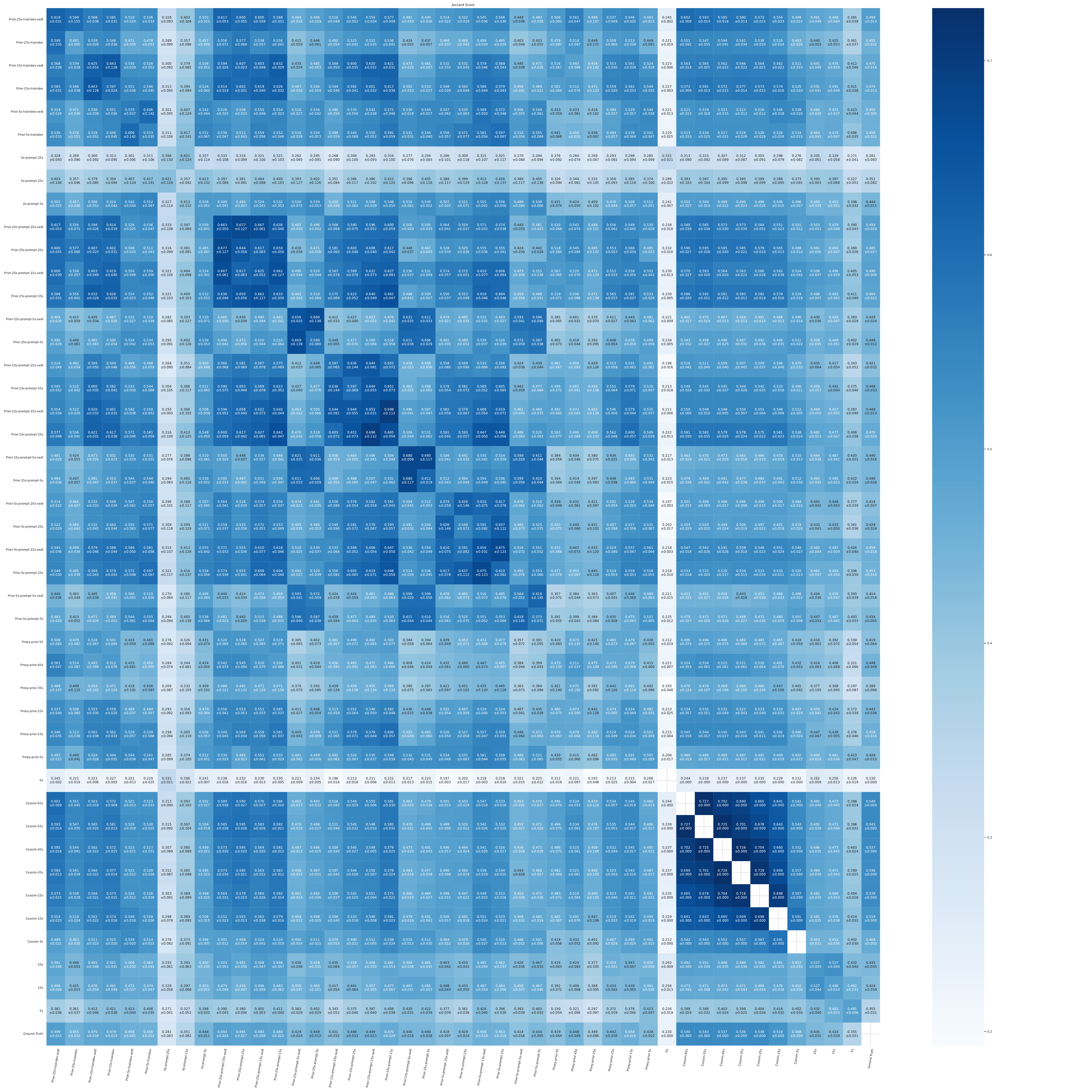}}
\caption{Jaccard Score similarities across \texttt{meta-llama/Llama-2-13b-chat-hf} experiments on SemEval}
\label{fig:js-llama13-semeval}
\end{figure*}

\begin{figure*}[htbp] 
\centerline{\includegraphics[scale=0.18]{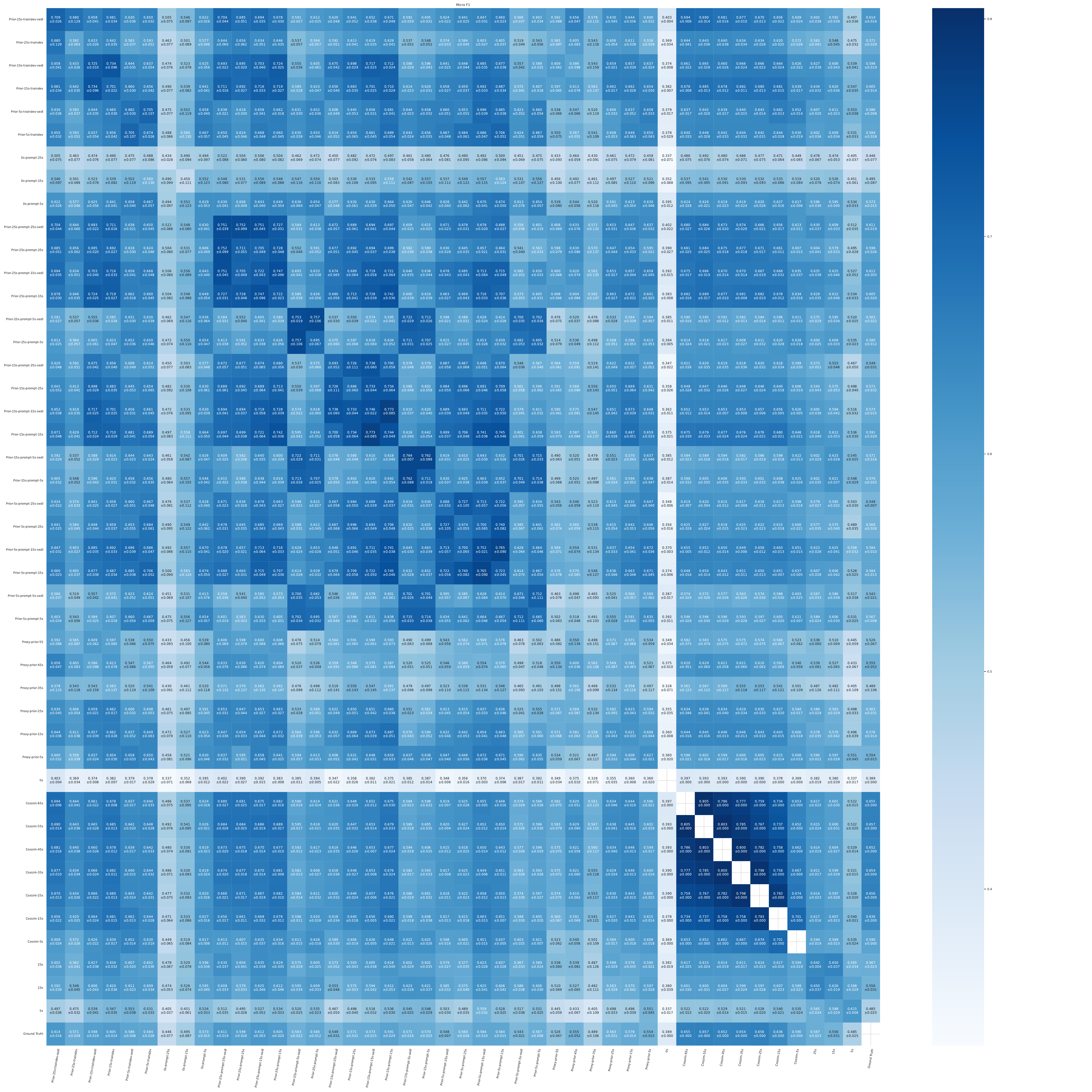}}
\caption{Micro F1 similarities across \texttt{meta-llama/Llama-2-13b-chat-hf} experiments on SemEval}
\label{fig:mic-llama13-semeval}
\end{figure*}

\begin{figure*}[htbp] 
\centerline{\includegraphics[scale=0.2]{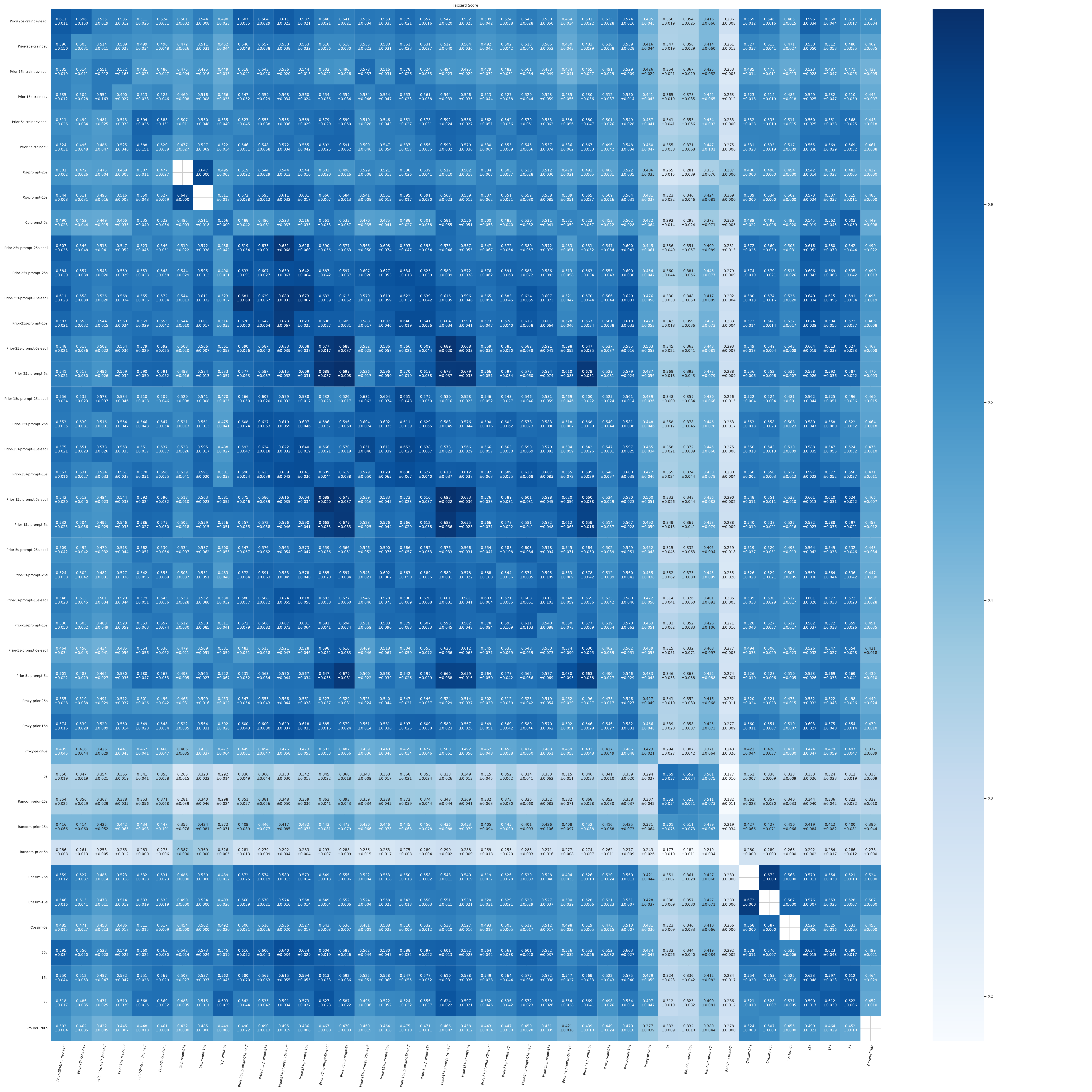}}
\caption{Jaccard Score similarities across \texttt{meta-llama/Llama-2-70b-chat-hf} experiments on SemEval}
\label{fig:js-llama70-semeval}
\end{figure*}

\begin{figure*}[htbp] 
\centerline{\includegraphics[scale=0.2]{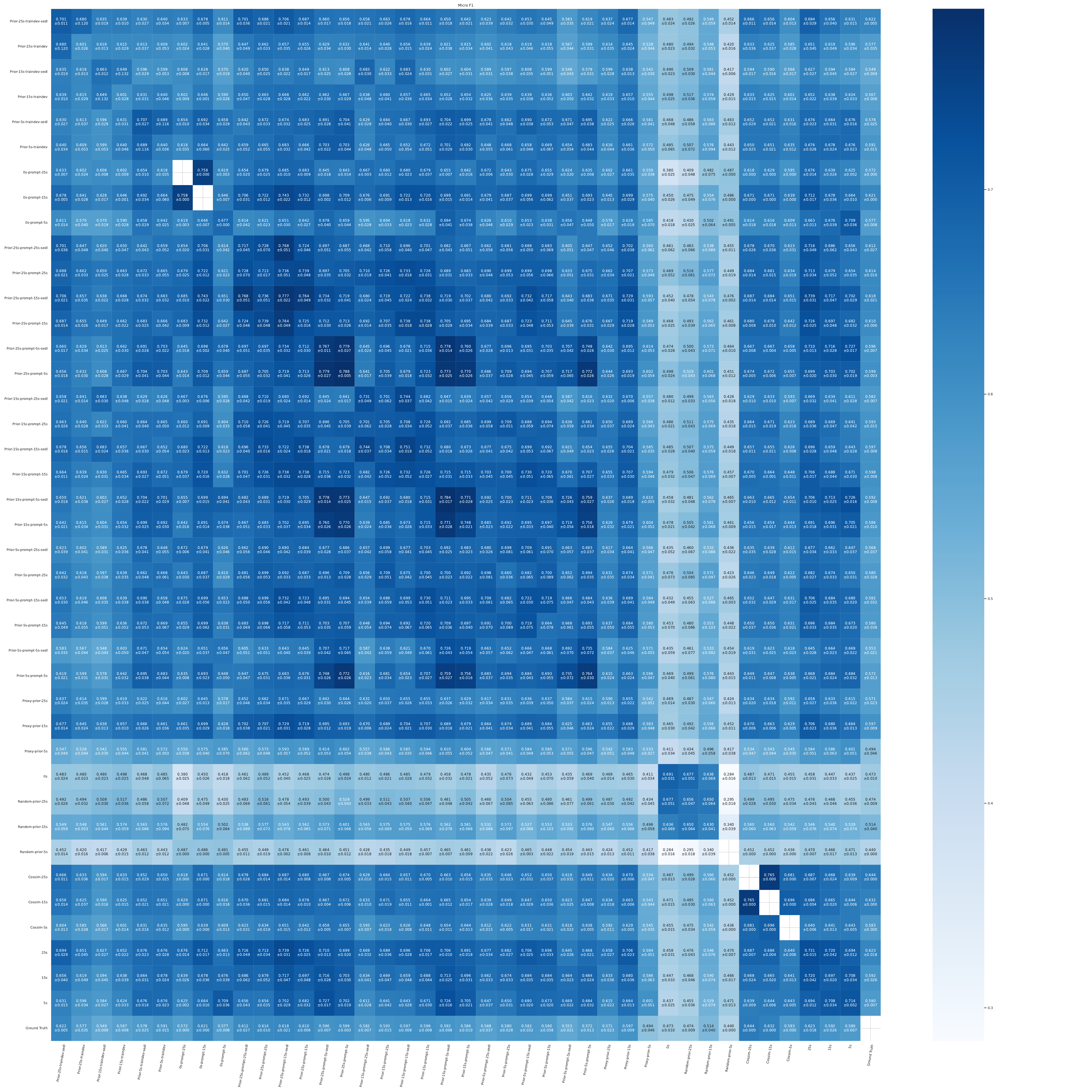}}
\caption{Micro F1 similarities across \texttt{meta-llama/Llama-2-70b-chat-hf} experiments on SemEval}
\label{fig:mic-llama70-semeval}
\end{figure*}

%% file: goemotions-results.tex
\begin{figure*}[htbp] 
\centerline{\includegraphics[scale=0.2]{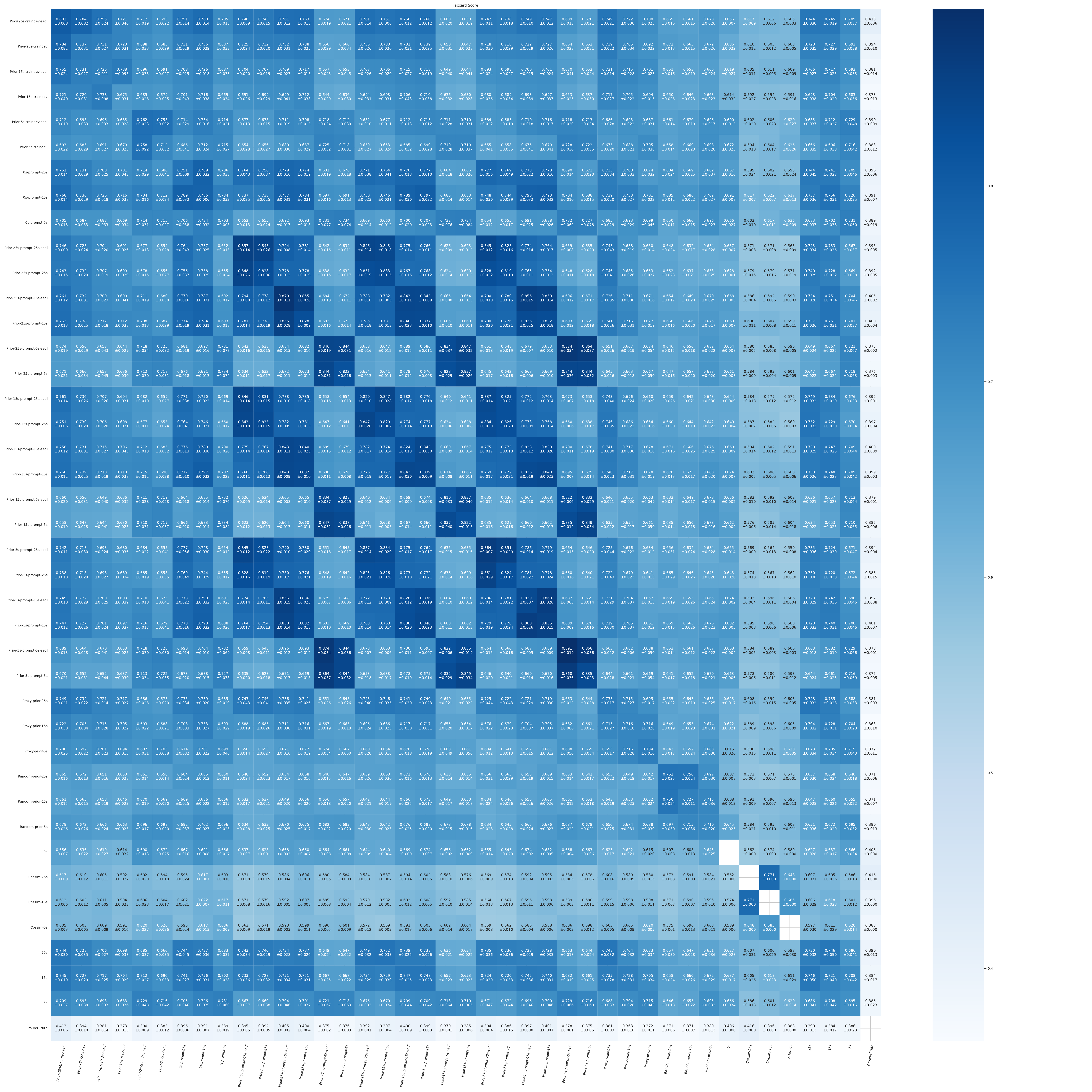}}
\caption{Jaccard Score similarities across \texttt{gpt-3.5-turbo} experiments on GoEmotions}
\label{fig:js-gpt-goemotions}
\end{figure*}

\begin{figure*}[htbp] 
\centerline{\includegraphics[scale=0.2]{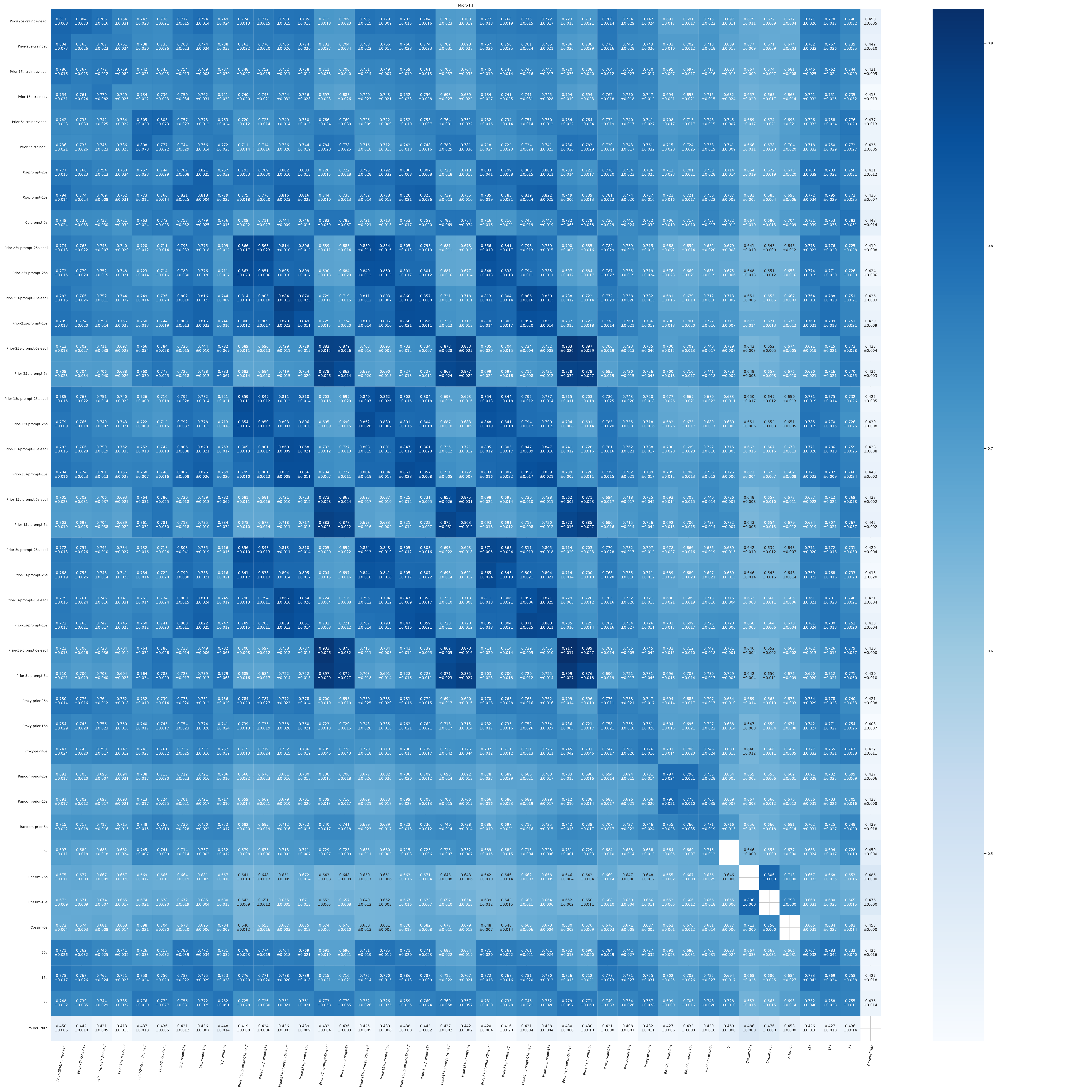}}
\caption{Micro F1 similarities across \texttt{gpt-3.5-turbo} experiments on GoEmotions}
\label{fig:mic-gpt-goemotions}
\end{figure*}

\begin{figure*}[htbp] 
\centerline{\includegraphics[scale=0.22]{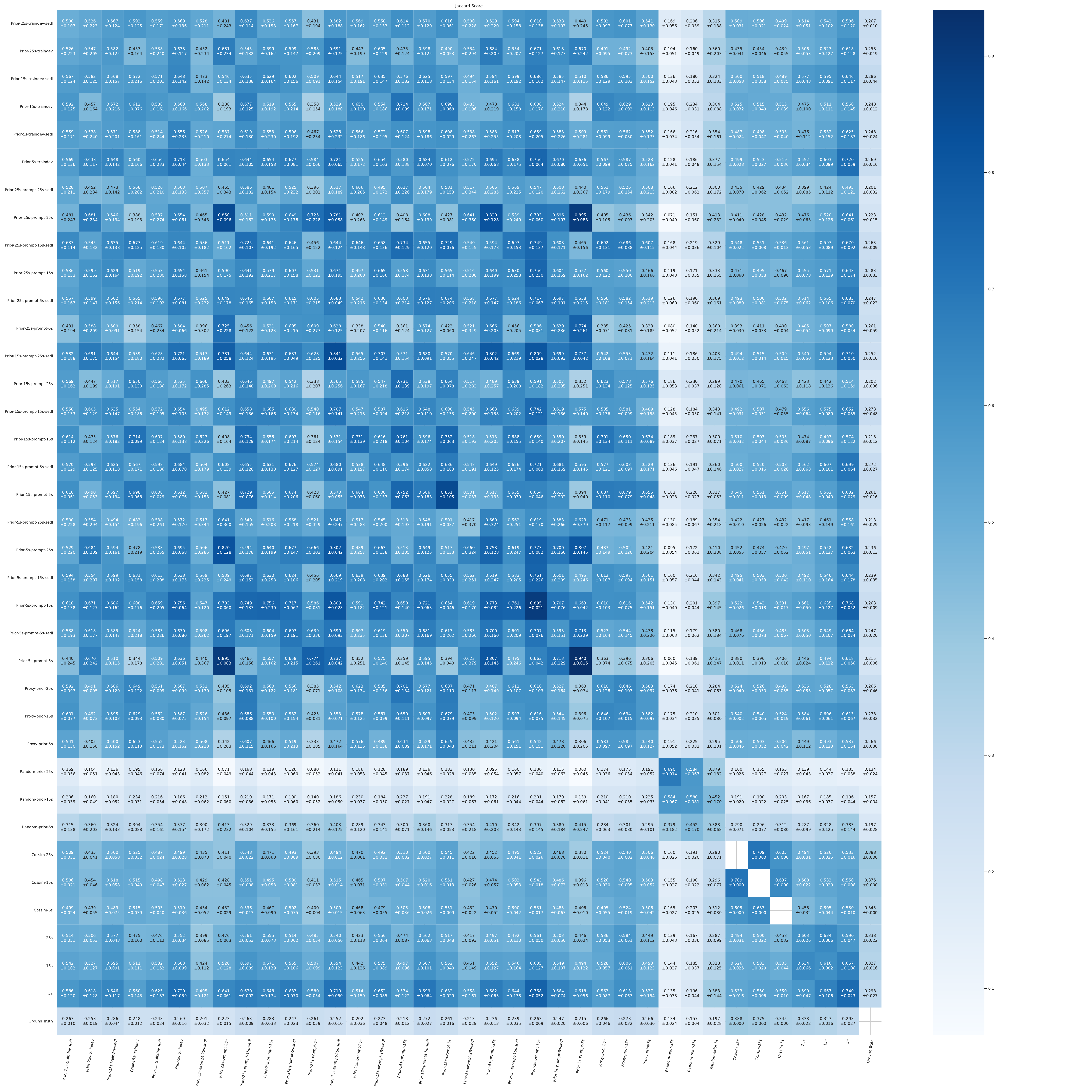}}
\caption{Jaccard Score similarities across \texttt{google/gemma-7b} experiments on GoEmotions}
\label{fig:js-gemma-goemotions}
\end{figure*}

\begin{figure*}[htbp] 
\centerline{\includegraphics[scale=0.22]{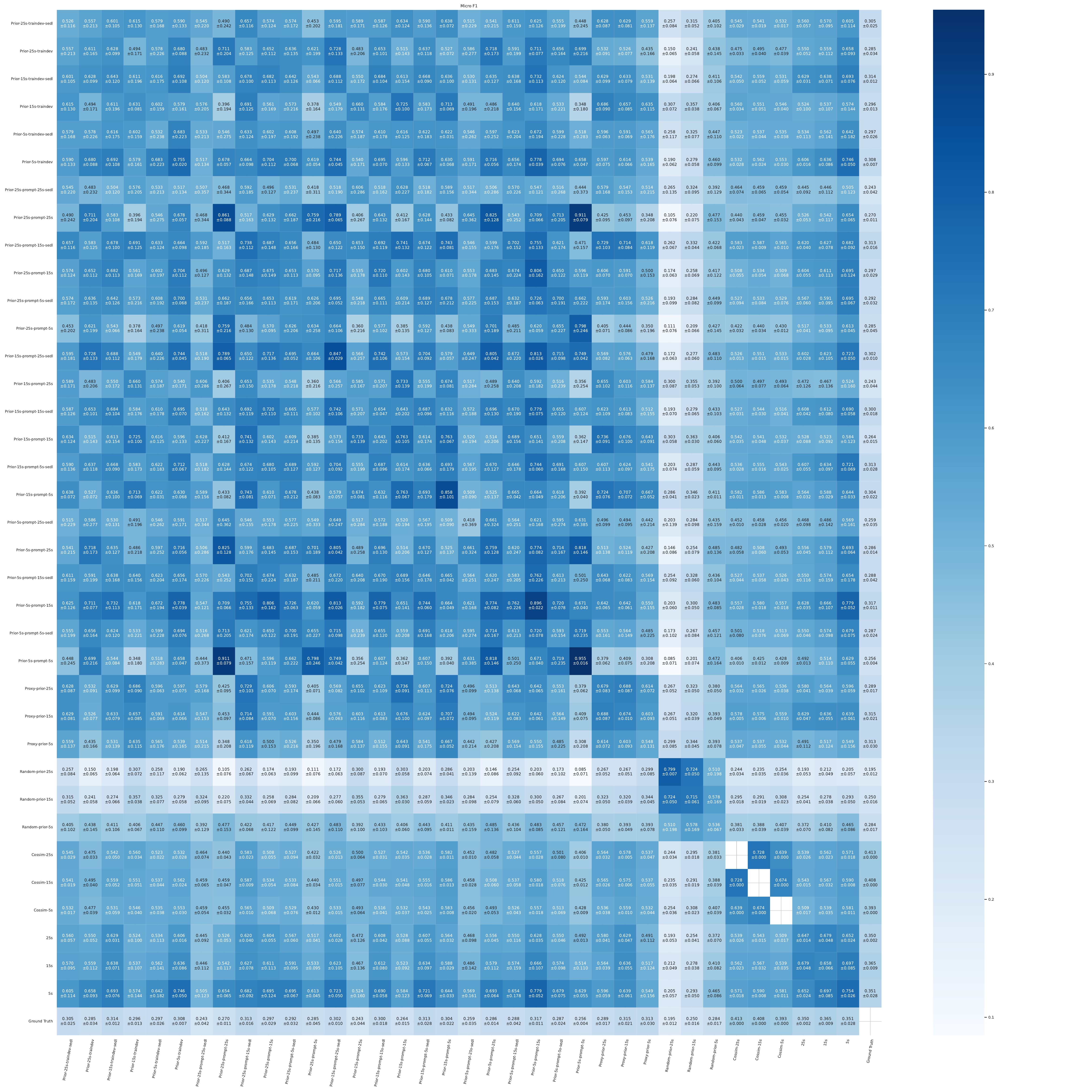}}
\caption{Micro F1 similarities across \texttt{google/gemma-7b} experiments on GoEmotions}
\label{fig:mic-gemma-goemotions}
\end{figure*}

\begin{figure*}[htbp] 
\centerline{\includegraphics[scale=0.22]{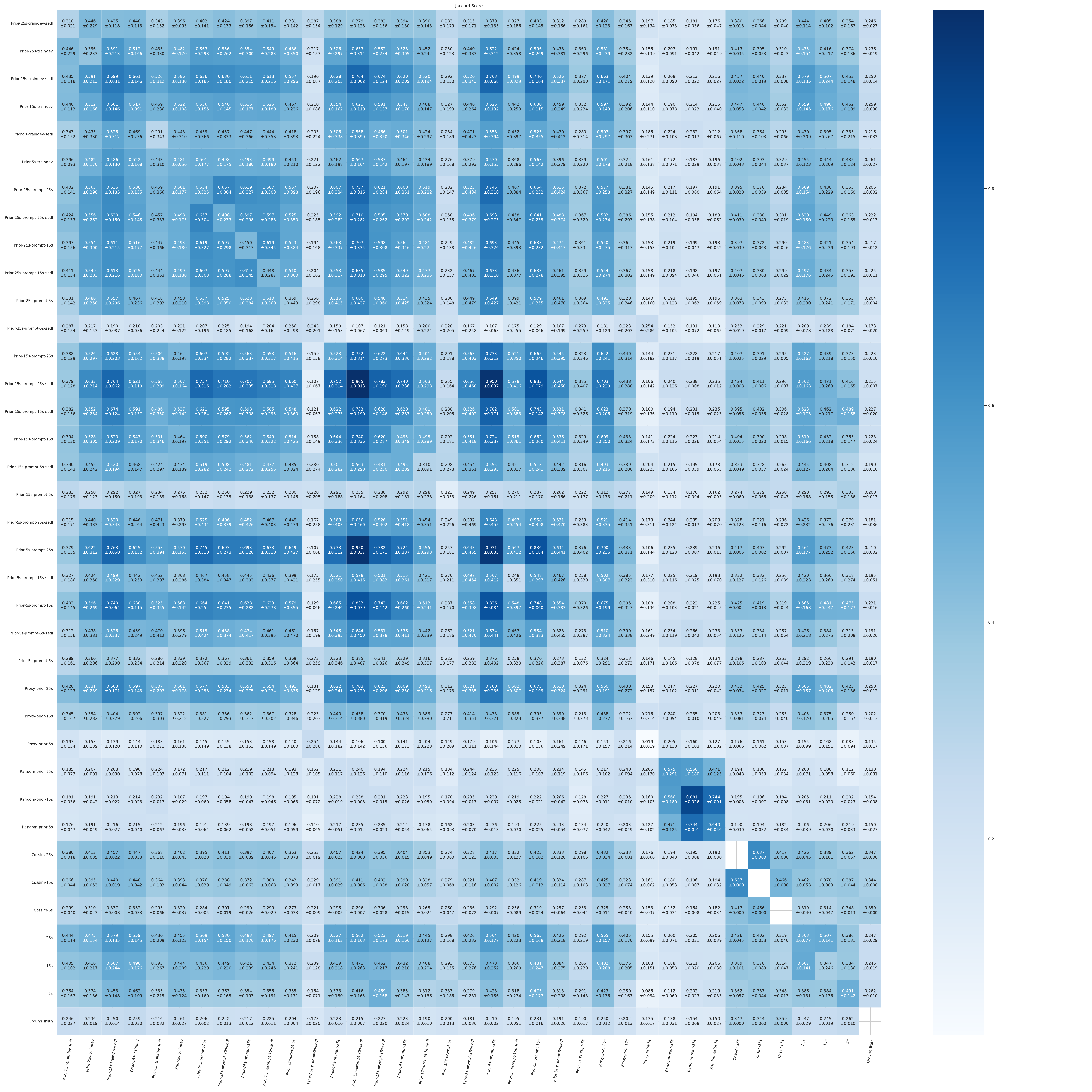}}
\caption{Jaccard Score similarities across \texttt{allenai/olmo-7B} experiments on GoEmotions}
\label{fig:js-olmo-goemotions}
\end{figure*}

\begin{figure*}[htbp] 
\centerline{\includegraphics[scale=0.2]{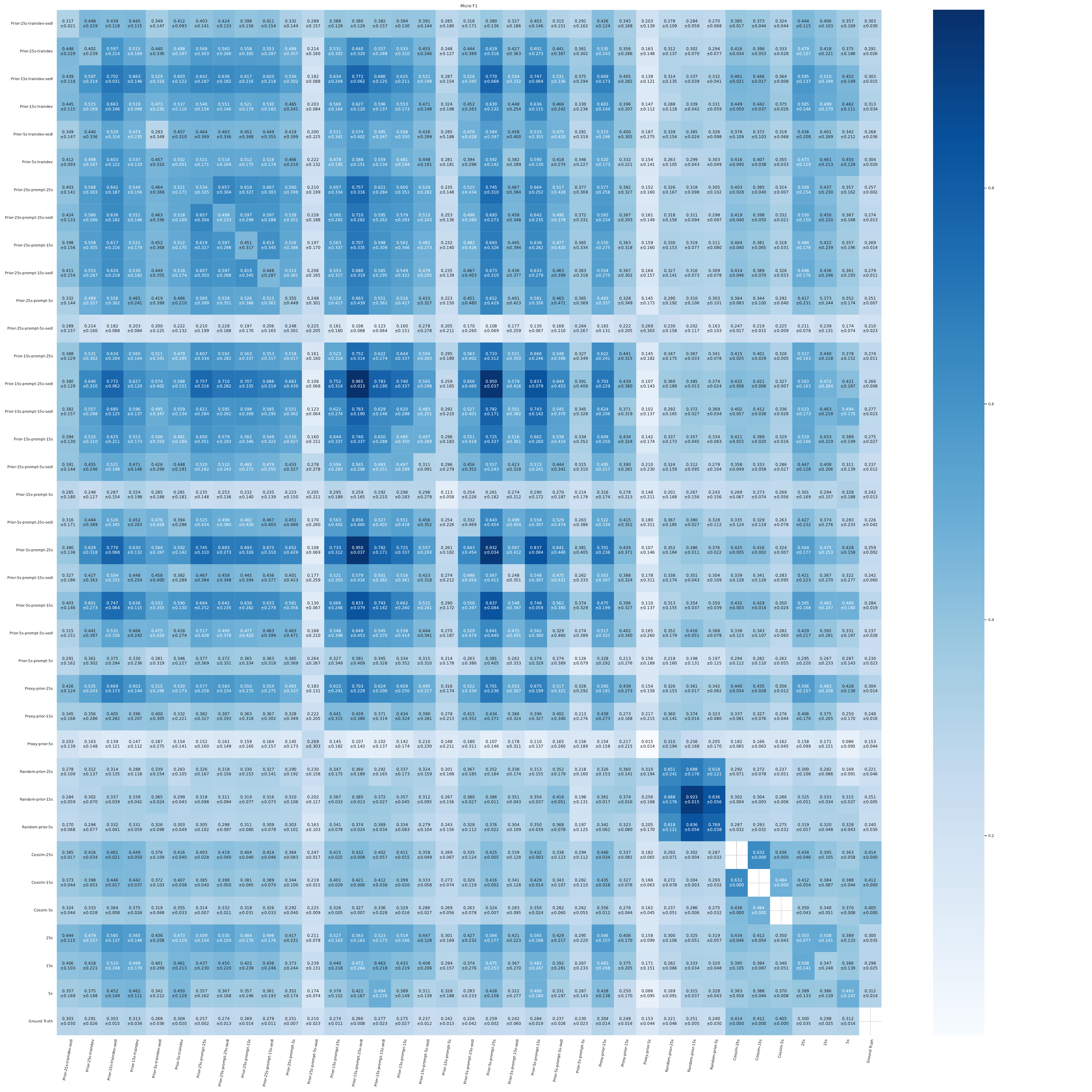}}
\caption{Micro F1 similarities across \texttt{allenai/olmo-7B} experiments on GoEmotions}
\label{fig:mic-olmo-goemotions}
\end{figure*}

\begin{figure*}[htbp] 
\centerline{\includegraphics[scale=0.2]{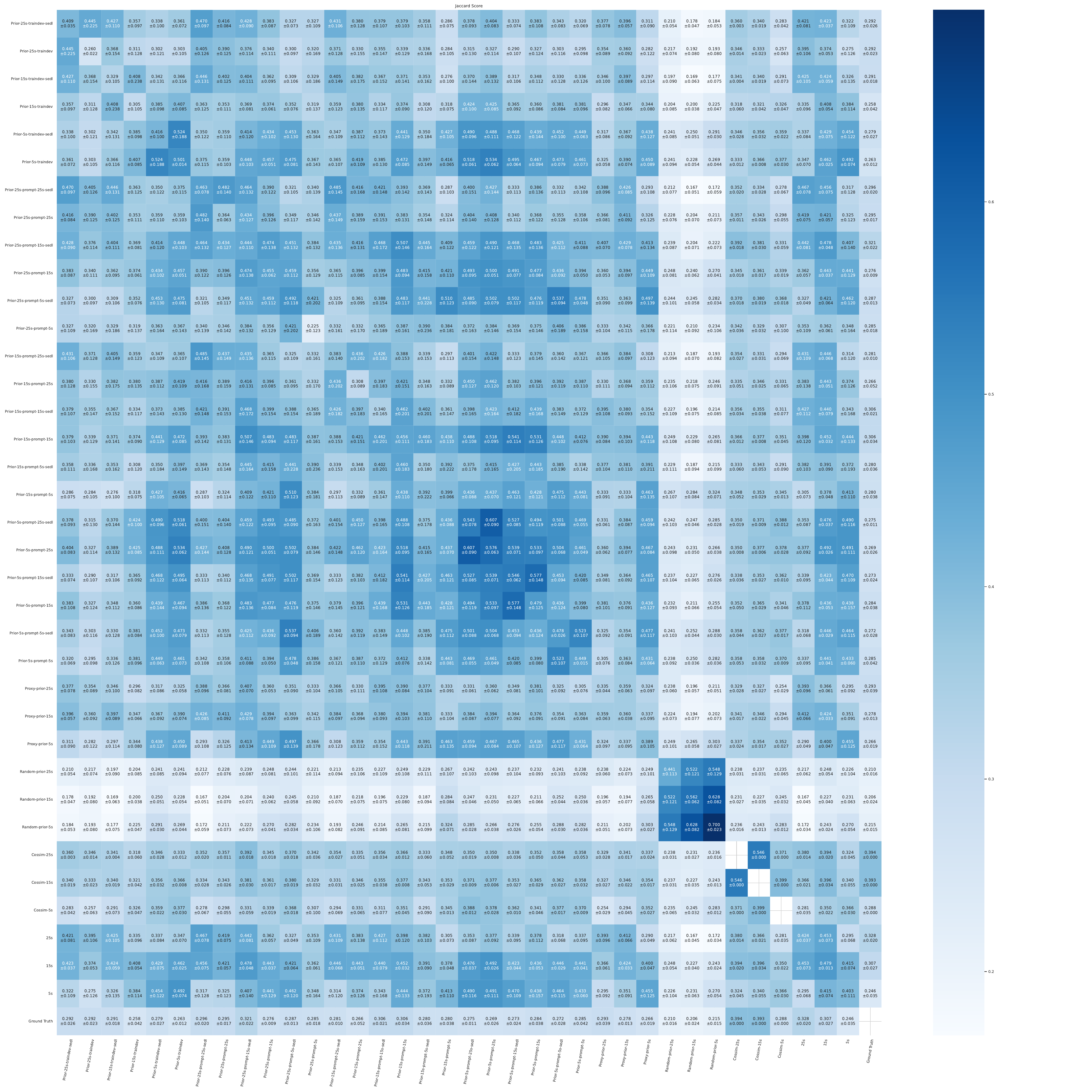}}
\caption{Jaccard Score similarities across \texttt{meta-llama/Llama-2-13b-chat-hf} experiments on GoEmotions}
\label{fig:js-llama13-goemotions}
\end{figure*}

\begin{figure*}[htbp] 
\centerline{\includegraphics[scale=0.2]{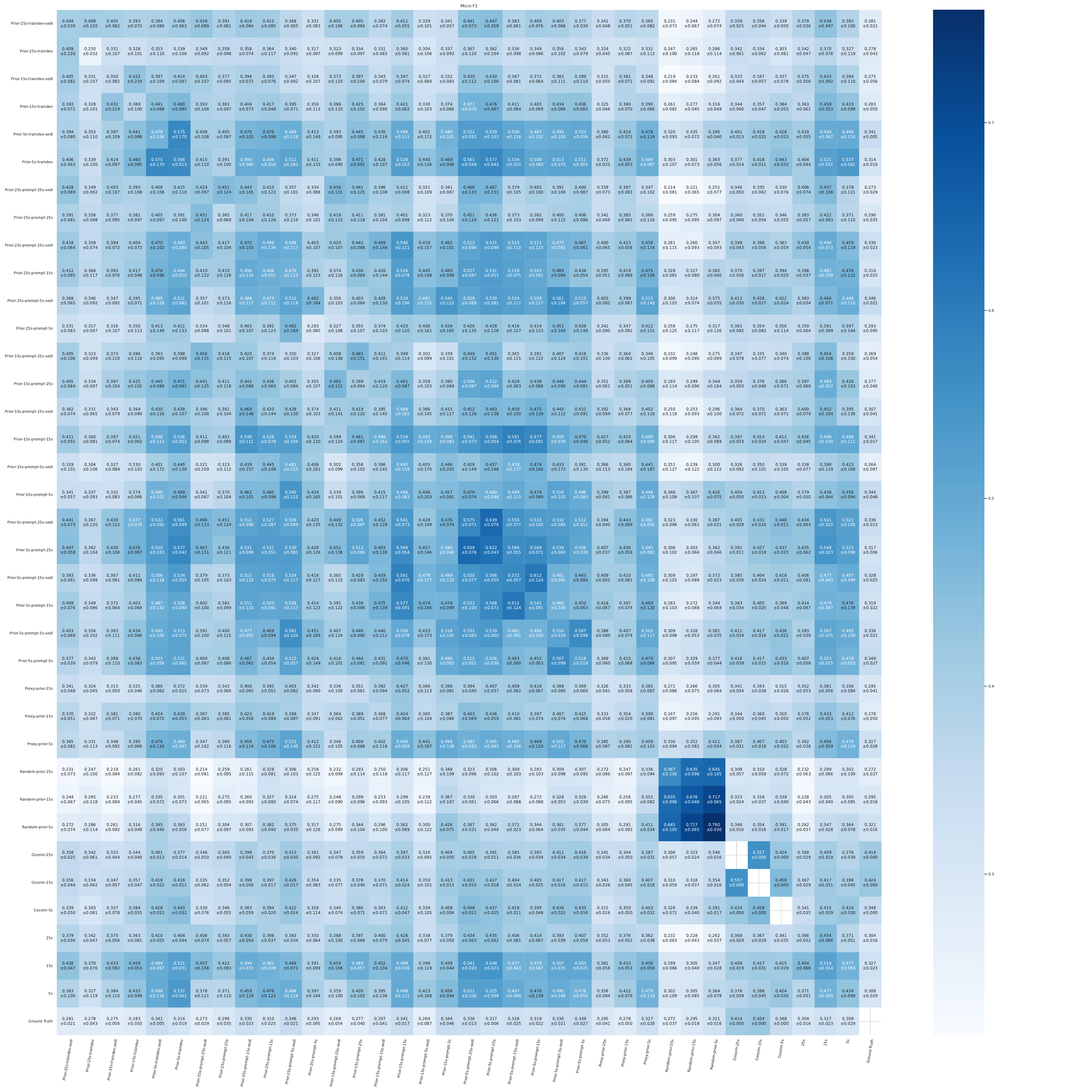}}
\caption{Micro F1 similarities across \texttt{meta-llama/Llama-2-13b-chat-hf} experiments on GoEmotions}
\label{fig:mic-llama13-goemotions}
\end{figure*}

\begin{figure*}[htbp] 
\centerline{\includegraphics[scale=0.4]{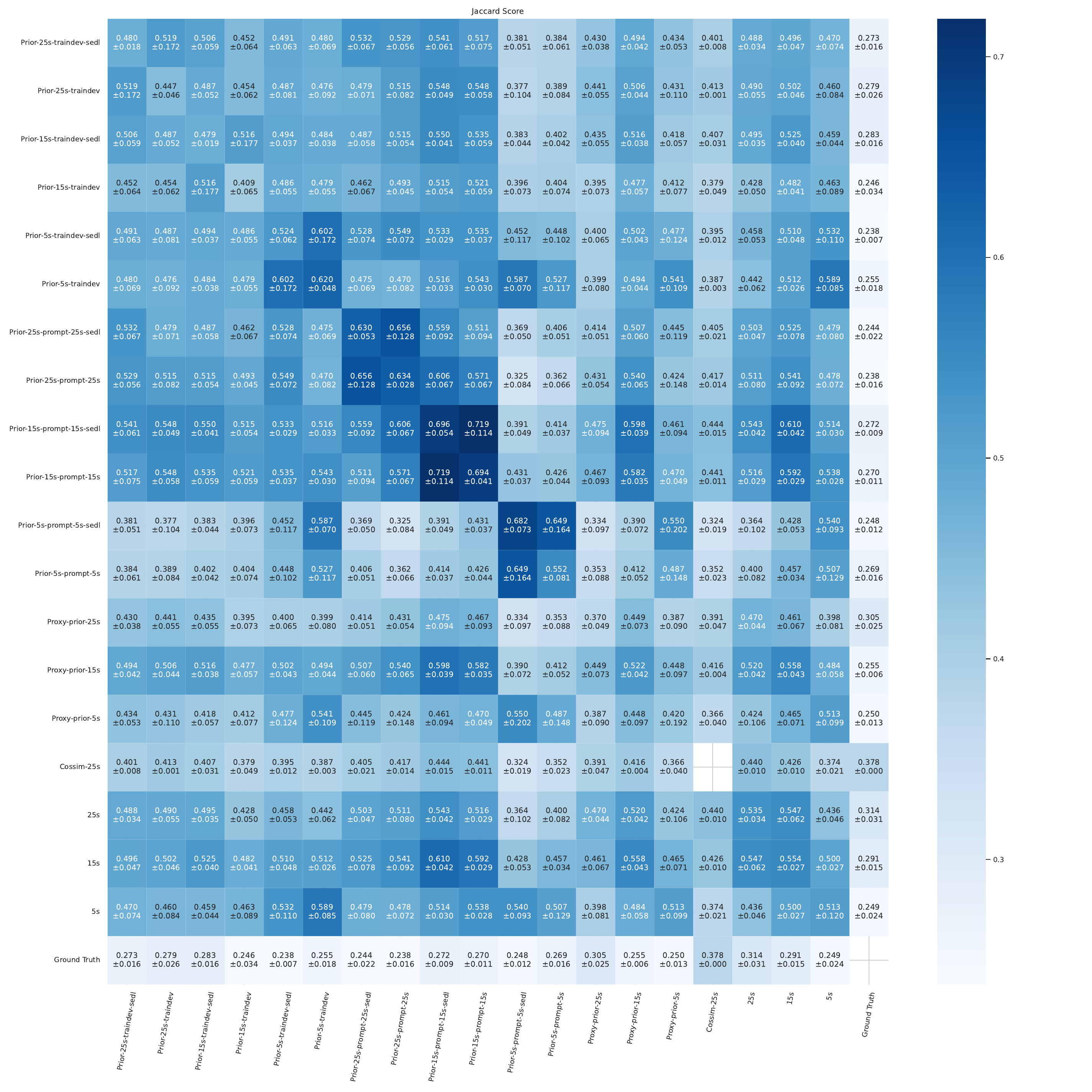}}
\caption{Jaccard Score similarities across \texttt{meta-llama/Llama-2-70b-chat-hf} experiments on GoEmotions}
\label{fig:js-llama70-goemotions}
\end{figure*}

\begin{figure*}[htbp] 
\centerline{\includegraphics[scale=0.4]{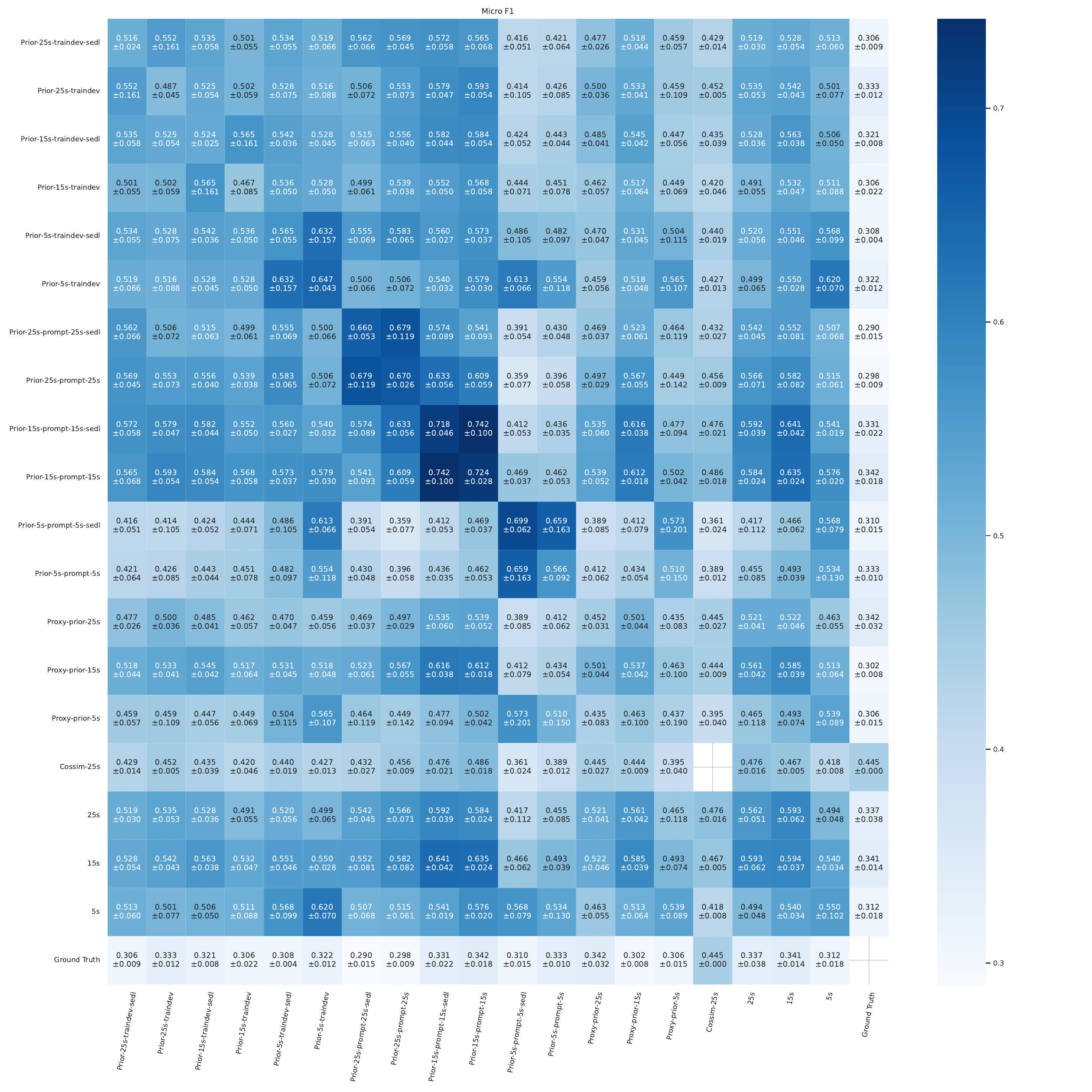}}
\caption{Micro F1 similarities across \texttt{meta-llama/Llama-2-70b-chat-hf} experiments on GoEmotions}
\label{fig:mic-llama70-goemotions}
\end{figure*}

%% file: conference_101719.bbl
\begin{thebibliography}{10}
\providecommand{\url}[1]{#1}
\csname url@samestyle\endcsname
\providecommand{\newblock}{\relax}
\providecommand{\bibinfo}[2]{#2}
\providecommand{\BIBentrySTDinterwordspacing}{\spaceskip=0pt\relax}
\providecommand{\BIBentryALTinterwordstretchfactor}{4}
\providecommand{\BIBentryALTinterwordspacing}{\spaceskip=\fontdimen2\font plus
\BIBentryALTinterwordstretchfactor\fontdimen3\font minus \fontdimen4\font\relax}
\providecommand{\BIBforeignlanguage}[2]{{%
\expandafter\ifx\csname l@#1\endcsname\relax
\typeout{** WARNING: IEEEtran.bst: No hyphenation pattern has been}%
\typeout{** loaded for the language `#1'. Using the pattern for}%
\typeout{** the default language instead.}%
\else
\language=\csname l@#1\endcsname
\fi
#2}}
\providecommand{\BIBdecl}{\relax}
\BIBdecl

\bibitem{radford2019language}
A.~Radford, J.~Wu, R.~Child, D.~Luan, D.~Amodei, I.~Sutskever \emph{et~al.}, ``Language models are unsupervised multitask learners,'' \emph{OpenAI blog}, vol.~1, no.~8, p.~9.

\bibitem{ouyangTrainingLanguageModels2022}
L.~Ouyang, J.~Wu, X.~Jiang, D.~Almeida, C.~L. Wainwright, P.~Mishkin, C.~Zhang, S.~Agarwal, K.~Slama, A.~Ray \emph{et~al.}, ``Training language models to follow instructions with human feedback.''

\bibitem{touvronLlamaOpenFoundation2023}
H.~Touvron, L.~Martin, K.~Stone, P.~Albert, A.~Almahairi, Y.~Babaei, N.~Bashlykov, S.~Batra, P.~Bhargava, and S.~Bhosale, ``Llama 2: {{Open}} foundation and fine-tuned chat models.''

\bibitem{Touvron2023a}
H.~Touvron, T.~Lavril, G.~Izacard, X.~Martinet, M.-A. Lachaux, T.~Lacroix, B.~Rozière, N.~Goyal, E.~Hambro, and F.~Azhar, ``Llama: {{Open}} and efficient foundation language models.''

\bibitem{zheng2023judging}
L.~Zheng, W.-L. Chiang, Y.~Sheng, S.~Zhuang, Z.~Wu, Y.~Zhuang, Z.~Lin, Z.~Li, D.~Li, E.~Xing \emph{et~al.}, ``Judging {{LLM-as-a-judge}} with {{MT-Bench}} and chatbot arena.''

\bibitem{brownLanguageModelsAre2020}
T.~Brown, B.~Mann, N.~Ryder, M.~Subbiah, J.~D. Kaplan, P.~Dhariwal, A.~Neelakantan, P.~Shyam, G.~Sastry, and A.~Askell, ``Language models are few-shot learners,'' \emph{Advances in neural information processing systems}, vol.~33, pp. 1877--1901.

\bibitem{achiam2023gpt}
J.~Achiam, S.~Adler, S.~Agarwal, L.~Ahmad, I.~Akkaya, F.~L. Aleman, D.~Almeida, J.~Altenschmidt, S.~Altman, S.~Anadkat \emph{et~al.}, ``Gpt-4 technical report.''

\bibitem{ziems2023can}
C.~Ziems, O.~Shaikh, Z.~Zhang, W.~Held, J.~Chen, and D.~Yang, ``Can large language models transform computational social science?'' pp. 1--53.

\bibitem{kim2024health}
Y.~Kim, X.~Xu, D.~McDuff, C.~Breazeal, and H.~W. Park, ``Health-llm: Large language models for health prediction via wearable sensor data,'' \emph{arXiv preprint arXiv:2401.06866}, 2024.

\bibitem{bakker2022fine}
M.~Bakker, M.~Chadwick, H.~Sheahan, M.~Tessler, L.~Campbell-Gillingham, J.~Balaguer, N.~McAleese, A.~Glaese, J.~Aslanides, M.~Botvinick \emph{et~al.}, ``Fine-tuning language models to find agreement among humans with diverse preferences,'' \emph{Advances in Neural Information Processing Systems}, vol.~35, pp. 38\,176--38\,189, 2022.

\bibitem{wei2022emergent}
J.~Wei, Y.~Tay, R.~Bommasani, C.~Raffel, B.~Zoph, S.~Borgeaud, D.~Yogatama, M.~Bosma, D.~Zhou, D.~Metzler \emph{et~al.}, ``Emergent abilities of large language models.''

\bibitem{kossen2023context}
J.~Kossen, T.~Rainforth, and Y.~Gal, ``In-context learning in large language models learns label relationships but is not conventional learning.''

\bibitem{chanDataDistributionalProperties2022}
S.~C. Chan, A.~Santoro, A.~K. Lampinen, J.~X. Wang, A.~K. Singh, P.~H. Richemond, J.~McClelland, and F.~Hill, ``Data distributional properties drive emergent in-context learning in transformers,'' in \emph{Advances in {{Neural Information Processing Systems}}}.

\bibitem{min2022rethinking}
S.~Min, X.~Lyu, A.~Holtzman, M.~Artetxe, M.~Lewis, H.~Hajishirzi, and L.~Zettlemoyer, ``Rethinking the role of demonstrations: What makes in-context learning work?'' in \emph{Proceedings of the 2022 Conference on Empirical Methods in Natural Language Processing}, 2022, pp. 11\,048--11\,064.

\bibitem{xie2021explanation}
S.~M. Xie, A.~Raghunathan, P.~Liang, and T.~Ma, ``An explanation of in-context learning as implicit bayesian inference,'' in \emph{International Conference on Learning Representations}, 2021.

\bibitem{wei2023larger}
J.~Wei, J.~Wei, Y.~Tay, D.~Tran, A.~Webson, Y.~Lu, X.~Chen, H.~Liu, D.~Huang, D.~Zhou \emph{et~al.}, ``Larger language models do in-context learning differently.''

\bibitem{pan2023context}
J.~Pan, T.~Gao, H.~Chen, and D.~Chen, ``What in-context learning" learns" in-context: {{Disentangling}} task recognition and task learning.''

\bibitem{demszkyGoEmotionsDatasetFinegrained2020}
D.~Demszky, D.~Movshovitz-Attias, J.~Ko, A.~Cowen, G.~Nemade, and S.~Ravi, ``{{GoEmotions}}: {{A}} dataset of fine-grained emotions.''

\bibitem{mohammad2018semeval}
S.~Mohammad, F.~Bravo-Marquez, M.~Salameh, and S.~Kiritchenko, ``Semeval-2018 task 1: {{Affect}} in tweets,'' in \emph{Proceedings of the 12th International Workshop on Semantic Evaluation}, pp. 1--17.

\bibitem{chochlakisLeveragingLabelCorrelations2023}
G.~Chochlakis, G.~Mahajan, S.~Baruah, K.~Burghardt, K.~Lerman, and S.~Narayanan, ``Leveraging label correlations in a multi-label setting: {{A}} case study in emotion,'' in \emph{{{ICASSP}} 2023-2023 {{IEEE International Conference}} on {{Acoustics}}, {{Speech}} and {{Signal Processing}} ({{ICASSP}})}.\hskip 1em plus 0.5em minus 0.4em\relax {IEEE}, pp. 1--5.

\bibitem{srivastava2022beyond}
A.~Srivastava, A.~Rastogi, A.~Rao, A.~A.~M. Shoeb, A.~Abid, A.~Fisch, A.~R. Brown, A.~Santoro, A.~Gupta, A.~Garriga-Alonso \emph{et~al.}, ``Beyond the imitation game: {{Quantifying}} and extrapolating the capabilities of language models.''

\bibitem{liu2022makes}
J.~Liu, D.~Shen, Y.~Zhang, B.~Dolan, L.~Carin, and W.~Chen, ``What makes good in-context examples for gpt-3?'' \emph{DeeLIO 2022}, p. 100, 2022.

\bibitem{rubin2022learning}
O.~Rubin, J.~Herzig, and J.~Berant, ``Learning to retrieve prompts for in-context learning,'' in \emph{Proceedings of the 2022 Conference of the North American Chapter of the Association for Computational Linguistics: Human Language Technologies}, 2022, pp. 2655--2671.

\bibitem{weiChainThoughtPrompting2022}
J.~Wei, X.~Wang, D.~Schuurmans, M.~Bosma, E.~Chi, Q.~Le, and D.~Zhou, ``Chain of thought prompting elicits reasoning in large language models.''

\bibitem{yao2024tree}
``Tree of thoughts: Deliberate problem solving with large language models,'' in \emph{Advances in Neural Information Processing Systems}, 2024.

\bibitem{zhao2021calibrate}
Z.~Zhao, E.~Wallace, S.~Feng, D.~Klein, and S.~Singh, ``Calibrate before use: Improving few-shot performance of language models,'' in \emph{International conference on machine learning}.\hskip 1em plus 0.5em minus 0.4em\relax PMLR, 2021, pp. 12\,697--12\,706.

\bibitem{santurkarWhoseOpinionsLanguage2023a}
S.~Santurkar, E.~Durmus, F.~Ladhak, C.~Lee, P.~Liang, and T.~Hashimoto, ``Whose opinions do language models reflect?''

\bibitem{yongsatianchot2023investigating}
N.~Yongsatianchot, P.~G. Torshizi, and S.~Marsella, ``Investigating large language models' perception of emotion using appraisal theory.''

\bibitem{hartmann2023political}
J.~Hartmann, J.~Schwenzow, and M.~Witte, ``The political ideology of conversational ai: Converging evidence on chatgpt’s pro-environmental, left-libertarian orientation,'' \emph{Left-Libertarian Orientation (January 1, 2023)}, 2023.

\bibitem{zhao2023group}
S.~Zhao, J.~Dang, and A.~Grover, ``Group preference optimization: Few-shot alignment of large language models,'' \emph{arXiv preprint arXiv:2310.11523}, 2023.

\bibitem{he2024whose}
Z.~He, S.~Guo, A.~Rao, and K.~Lerman, ``Whose emotions and moral sentiments do language models reflect?'' \emph{arXiv preprint arXiv:2402.11114}, 2024.

\bibitem{zou2023universal}
A.~Zou, Z.~Wang, J.~Z. Kolter, and M.~Fredrikson, ``Universal and transferable adversarial attacks on aligned language models,'' \emph{arXiv preprint arXiv:2307.15043}, 2023.

\bibitem{cowenSelfreportCaptures272017}
A.~S. Cowen and D.~Keltner, ``Self-report captures 27 distinct categories of emotion bridged by continuous gradients,'' \emph{Proceedings of the national academy of sciences}, vol. 114, no.~38, pp. E7900--E7909.

\bibitem{wolf-etal-2020-transformers}
\BIBentryALTinterwordspacing
T.~Wolf, L.~Debut, V.~Sanh, J.~Chaumond, C.~Delangue, A.~Moi, P.~Cistac, T.~Rault, R.~Louf, M.~Funtowicz, J.~Davison, S.~Shleifer, P.~von Platen, C.~Ma, Y.~Jernite, J.~Plu, C.~Xu, T.~L. Scao, S.~Gugger, M.~Drame, Q.~Lhoest, and A.~M. Rush, ``Transformers: State-of-the-art natural language processing,'' in \emph{Proceedings of the 2020 Conference on Empirical Methods in Natural Language Processing: System Demonstrations}.\hskip 1em plus 0.5em minus 0.4em\relax Online: Association for Computational Linguistics, Oct. 2020, pp. 38--45. [Online]. Available: \url{https://www.aclweb.org/anthology/2020.emnlp-demos.6}
\BIBentrySTDinterwordspacing

\bibitem{song2020mpnet}
K.~Song, X.~Tan, T.~Qin, J.~Lu, and T.-Y. Liu, ``Mpnet: Masked and permuted pre-training for language understanding,'' \emph{Advances in neural information processing systems}, vol.~33, pp. 16\,857--16\,867, 2020.

\bibitem{caliskan2017semantics}
A.~Caliskan, J.~J. Bryson, and A.~Narayanan, ``Semantics derived automatically from language corpora contain human-like biases,'' \emph{Science}, vol. 356, no. 6334, pp. 183--186, 2017.

\bibitem{gonen2019lipstick}
H.~Gonen and Y.~Goldberg, ``Lipstick on a pig: Debiasing methods cover up systematic gender biases in word embeddings but do not remove them,'' \emph{arXiv preprint arXiv:1903.03862}, 2019.

\bibitem{ferrara2023should}
E.~Ferrara, ``Should chatgpt be biased? challenges and risks of bias in large language models,'' \emph{arXiv preprint arXiv:2304.03738}, 2023.

\bibitem{yang2023context}
D.~Yang, A.~Kommineni, M.~Alshehri, N.~Mohanty, V.~Modi, J.~Gratch, and S.~Narayanan, ``Context unlocks emotions: Text-based emotion classification dataset auditing with large language models,'' in \emph{2023 11th International Conference on Affective Computing and Intelligent Interaction (ACII)}.\hskip 1em plus 0.5em minus 0.4em\relax IEEE, 2023, pp. 1--8.

\bibitem{swayamdipta2020dataset}
S.~Swayamdipta, R.~Schwartz, N.~Lourie, Y.~Wang, H.~Hajishirzi, N.~A. Smith, and Y.~Choi, ``Dataset cartography: {{Mapping}} and diagnosing datasets with training dynamics.''

\bibitem{aroyo2015truth}
L.~Aroyo and C.~Welty, ``Truth is a lie: {{Crowd}} truth and the seven myths of human annotation,'' \emph{AI Magazine}, vol.~36, no.~1, pp. 15--24.

\bibitem{hu2021lora}
E.~J. Hu, P.~Wallis, Z.~Allen-Zhu, Y.~Li, S.~Wang, L.~Wang, W.~Chen \emph{et~al.}, ``Lora: Low-rank adaptation of large language models,'' in \emph{International Conference on Learning Representations}, 2021.

\end{thebibliography}
